\documentclass[journal]{IEEEtran}
\usepackage{microtype}
\usepackage{graphicx}
\usepackage{booktabs} 

\usepackage{times}
\usepackage{epsfig}
\usepackage{graphicx}
\usepackage{amsthm}
\usepackage{amsmath}
\usepackage{amssymb}
\usepackage{mathrsfs}

\usepackage[utf8]{inputenc} 
\usepackage[T1]{fontenc}    
\usepackage{url}            
\usepackage{booktabs}       
\usepackage{amsfonts}       
\usepackage{nicefrac}       
\usepackage{microtype}      

\usepackage{algorithm}
\usepackage[algo2e,ruled,vlined]{algorithm2e}
\usepackage{algorithmic}

\usepackage{graphicx} 
\usepackage{subcaption}
\usepackage{amsmath}
\usepackage{amsthm}
\usepackage{amssymb}
\usepackage{mathrsfs}
\usepackage{bm}
\usepackage{multirow}
\usepackage{multicol}
\newcommand{\algrule}[1][.2pt]{\par\vskip.5\baselineskip\hrule height #1\par\vskip.5\baselineskip}

\theoremstyle{plain}

\usepackage{breqn}

\newcounter{ass_counter}
\newcounter{thm_counter}
\newtheorem{theorem}[thm_counter]{Theorem}
\newtheorem{lemma}[thm_counter]{Lemma}

\newtheorem{assumption}[ass_counter]{Assumption}

\SetNlSty{}{}{}

\let\oldnl\nl
\newcommand\nonl{%
  \renewcommand{\nl}{\let\nl\oldnl}}

\def\1{{\bf{1}}}
\def\0{{\bf{0}}}

\def\u{{\bf u}}

\def\x{{\bf x}}

\newcommand{\beq}{\begin{equation}}
   \newcommand{\eeq}{\end{equation}}

\newcommand{\R}{\mathbb{R}}
\newcommand{\B}{\mathbb{B}}

\usepackage{hyperref}

\usepackage{cite}

\ifCLASSINFOpdf
\else
\fi
\begin{document}

\title{Learning Simple Thresholded Features with Sparse Support Recovery}
\author{Hongyu Xu*,~\IEEEmembership{Student Member,~IEEE},
        Zhangyang Wang*,~\IEEEmembership{Member,~IEEE},
        Haichuan Yang, \\
        Ding Liu,~\IEEEmembership{Member,~IEEE}, 
        and Ji Liu
\thanks{H. Xu is with the Department of Electrical and Computer Engineering, University of Maryland, College Park, MD, USA.}  
\thanks{E-mail: hyxu@umd.edu.}
\thanks{Z. Wang is with the Department of Computer Science and Engineering, Texas A\&M University, College Station, TX, USA.} 
\thanks{E-mail:  atlaswang@tamu.edu.}
\thanks{H. Yang and J. Liu are with the Department of Computer Science, University of Rochester, Rochester, NY, USA.}  
\thanks{E-mail: \{hyang36, jliu\}@cs.rochester.edu.}
\thanks{D. Liu is with the Beckman Institute, University of Illinois at Urbana-Champaign, Urbana, IL, USA.} 
\thanks{E-mail: dingliu2@illinois.edu.}
\thanks{*The first two authors contribute equally.}
}
\maketitle

\begin{abstract}
The thresholded feature has recently emerged as an extremely efficient, yet rough empirical approximation, of the time-consuming sparse coding inference process. Such an approximation has not yet been rigorously examined, and standard dictionaries often lead to non-optimal performance when used for computing thresholded features. In this paper, we first present two theoretical recovery guarantees for the thresholded feature to exactly recover the nonzero support of the sparse code. Motivated by them, we then formulate the \textit{Dictionary Learning for Thresholded Features} (DLTF) model, which learns an optimized dictionary for applying the thresholded feature. In particular, for the $(k, 2)$ norm involved, a novel proximal operator with log-linear time complexity $O(m\log m)$ is derived. We evaluate the performance of DLTF on a vast range of synthetic and real-data tasks, where DLTF demonstrates remarkable efficiency, effectiveness and robustness in all experiments. In addition, we briefly discuss the potential link between DLTF and deep learning building blocks.
\end{abstract}


%
\IEEEpeerreviewmaketitle

\section{Introduction}
%
%
%
%
Let $\Omega_k = \{\boldsymbol{z} \in \R^m: ||\boldsymbol{z}||_0 \le k\}$. For a data sample $\boldsymbol{x} \in \R^n$, the \textit{sparse coding} technique \cite{KSVD} aims to find the sparse code $\boldsymbol{z} \in \Omega_k$ to represent $\boldsymbol{x}$ compactly, i.e., $\boldsymbol{x} \approx W \boldsymbol{z}$, using a dictionary $W \in \R^{n \times m}$ =  $[\boldsymbol{w}_1, \boldsymbol{w}_2, ..., \boldsymbol{w}_m]$, where each atom $\boldsymbol{w}_i \in \R^n$ is assumed to have unit $\ell_2$-norm, $i = 1, ..., m$, to avoid scale ambiguity. With a properly designed or learned $W$, sparse coding is known to be powerful in numerous reconstruction or discriminative tasks such as signal sensing, classification and clustering~\cite{wang2016sparse,Xu_XZAC_ICPR2016,zhang2011sparse,cherian2012robust,Barthelemy_BLMMM_TSP2012,Ramirez_RS_TSP2012,Zhang_ZLCZY_TSP2016,Hyder_HM_TSP2010,Xu_2018,Zha_ZYTWC_KDE2014,Liu_LZTLYLM_ACMMM2016,Sun_SWLL_2018,liu2018alista} . One crucial drawback of sparse coding lies in its prohibitive cost of computing the sparse code at test time, which calls for iterative greedy or convex optimization algorithms \cite{blumensath2008iterative,Engan_EAH_SP2000}. This drawback limits the applicability of sparse codes in large-scale, high-dimensional problems, or when nearly real-time processing is desired.

Among a few fast sparse coding approximations, the simplest choice is arguably the \textit{thresholded feature} \cite{denil2012recklessly, fawzi2015dictionary,Mathilde_MDK_2012,Sprechmann_SLYBS_NIPS2013,Chen_CLWY_NIPS2018}: 
\begin{equation}
 \begin{aligned}
\bar{\boldsymbol{z}} := \text{max}_{k}(W^\top \boldsymbol{x}),
\label{feature}
  \end{aligned}
\end{equation}
 where $\text{max}_{k}$ retains $k$ largest-magnitude entries while setting others to zero\footnote{\cite{fawzi2015dictionary} exploited a ``soft'' version of the thresholded feature, which can be analyzed similarly.}. The threshold feature $\bar{\boldsymbol{z}}$ also belongs to $\Omega_k$, and is extremely efficient and easy to implement as it involves only a matrix-vector multiplication and a $\text{max}_{k}$ operation. \cite{coates2010analysis, coates2011importance} showed that such a simple encoding displays remarkable discriminative ability, and can often achieve comparable results to standard sparse coding, provided that the number of labeled samples and the dictionary size are large enough. \cite{fawzi2015dictionary} pointed out that the thresholded feature corresponds to an inexact approximation of sparse coding, where only one iteration of proximal gradient algorithm is unfolded \cite{wang2016learninga,wang2016learningb,wang2017doubly}. One may also notice the interesting resemblance of (\ref{feature}) to a standard linear fully connected layer
 plus neurons in deep learning \cite{ImageNet,wang2016d3}, on which we will discuss more later.

While dictionary learning~\cite{Rubinstein_RBE_TSP2010,mairal2012task,fawzi2015dictionary,Xu_XZC_BMVC2015,Xu_XZAC_WACV2016,Xu_XZAC_ICPR2016,Ramirez_RS_TSP2012,Rubinstein_RZE_TSP2010,Naderahmadian_NBT_TSP2016,Rubinstein_RE_TSP2014,Xu_2018,Bansal_BCW_NIPS2018} has been well developed for standard sparse coding \cite{KSVD}, the choice of $W$ remains barely explored for the thresholded feature. \cite{denil2012recklessly} used standard dictionaries, leading to a fairly rough approximation to the exact iterative solution, with sub-optimal results. \cite{fawzi2015dictionary} relied on supervised joint training to learn $W$, which is similar to learning a single-layer neural network classifier and does not generalize to unsupervised feature learning. Moreover, it is unclear how ``roughly'' $\bar{\boldsymbol{z}}$ approximates $\boldsymbol{z}$, and in what sense the former can be treated as a reliable substitute for the latter, leaving the effectiveness and robustness of threshold features under question.

This paper answers the above questions. Firstly, we show that under certain conditions, the thresholded feature $\bar{\boldsymbol{z}}$ exactly recovers the nonzero support of the sparse code $\boldsymbol{z}$. The support recovery is a core problem in sparse signal recovery \cite{yuan2016exact}: if the nonzero support set is correctly identified, $\boldsymbol{z}$ can be obtained using least squares method. Moreover, the support itself makes a useful feature in certain scenarios, such as quantization and hashing \cite{cherian2012robust}. 

Secondly, based on the theoretical guarantee forms, we derive the model and algorithm for \textit{Dictionary Learning for Thresholded Features} (DLTF), to learn the dictionary that is optimized for the thresholded feature. It is important to note that DLTF is not ``yet another'' way of \textit{standard dictionary learning}, whose inference relies on iterative sparse solvers. Instead, it is a new type of dictionary learning, specifically designed for using the thresholded feature (\ref{feature}) at inference, which prioritize (extreme) simplicity, efficiency, general usability and theoretical guarantees.

Last but not least, in particular, we derive a novel efficient $O(m\log m)$ algorithm for the $(k, 2)$ norm proximal subproblem. Both synthetic simulations and real-data experiments (\textit{i.e.} image clustering, denoising and unsupervised hashing) verify the competitive quantitative results and remarkable efficiency of applying thresholded features with DLTF dictionaries.

\section{Support Recovery Guarantees for Thresholded Features}

We present two support recovery guarantees, called the \textit{weak} and the \textit{strong} recovery guarantees respectively. The weak recovery guarantee depends on the \textit{mutual incoherence} \cite{donoho2006compressed}, and the strong recovery guarantee takes advantage of the \textit{Restricted Isometry Property} (RIP) \cite{candes2005decoding}. The two guarantees are called ``weak'' and ``strong'' respectively because of different sample complexity requirements: based on the results of \cite{zhang2011sparse}, to uniformly recover $\boldsymbol{z} \in \Omega_k$, the former requires $\mathcal{O}(k^2 \ln m)$ samples, while the latter gives rise to a lower sample complexity of $\mathcal{O}(k \ln m)$.

The two guarantees can be derived using the classical techniques adopted by the compressive sensing theory  \cite{foucart2013mathematical} and iterative thresholding algorithm \cite{blumensath2008iterative,Gribonval_GRSV_2008,Golbabaee_GV_ICASSP2008}. \textit{Full proof is detailed in the appendix~\ref{proof_Theorem1} and~\ref{proof_Theorem2}}. It is noted that our goal here is not to provide a tighter bound than existing literature (\textit{i.e.}, Theorem 5.16 of \cite{foucart2013mathematical}), but rather to illustrate what factors or quantities will affect the support recovery in the special case of thresholded features. We will further discuss how these guarantees motivate the proposed DLTF model in next section.


\subsection{The Weak Recovery Guarantee}

For a simple ``noiseless'' model: $\boldsymbol{x} = W \boldsymbol{z}$, without loss of generality, we assume that the non-zero entries of $\boldsymbol{z}$ are sorted by absolute magnitude in a decreasing order: $|z_1| \ge |z_2|... \ge |z_{k}|$. We denote by supp($\boldsymbol{z}$) $\in \B^m$ the sparse support of $\boldsymbol{z}$, i.e., its $i$-th entry is 1 if $z_i$ is nonzero. We define the mutual incoherence \cite{donoho2006compressed} of $W$: 
$\mu_W= \underset{i \neq j}{\text{max}} |\langle \boldsymbol{w}_i, \boldsymbol{w}_j \rangle|$. 


\begin{theorem}\label{thm1}
If the sufficient condition $k\mu_W \leq {|z_{k}| \over {2|z_1|}}$ holds, then
$\text{supp}(\boldsymbol{z})$ = $\text{supp}(\bar{\boldsymbol{z}})$, where $\bar{\boldsymbol{z}} := \text{max}_{k}(W^\top \boldsymbol{x}),$ .
\end{theorem}



A special case when $\boldsymbol{z}$ $\in \{0,1\}^{k}$ follows immediately (the $\text{sgn indicator}$ function is defined to yield output 1 when the input is nonzero, and 0 elsewhere):  

\textbf{Corollary 1.1} Assume $\boldsymbol{x} = W \boldsymbol{z}$, $\boldsymbol{z} \in \B^m$ and $||\boldsymbol{z}||_0 \le k$. If $\mu_W \leq {{1} \over {2k}}$, then $\boldsymbol{z}$ = $\text{sgn}(\bar{\boldsymbol{z}})$.

Theorem \ref{thm1} can be further (loosely) extended to the noisy case, when $\bm{x}$ is corrupted by the noise $\boldsymbol{e}$: $\boldsymbol{x} = W \boldsymbol{z} + \boldsymbol{e}$. Denote the mutual coherence between the dictionary and the noise: $\mu_e = \underset{i}{\text{max}} |\langle \boldsymbol{w}_i, \boldsymbol{e} \rangle|$, $\boldsymbol{e}$ may follow any statistical distribution only if $\mu_e$ can be properly bounded. 

\textbf{Corollary 1.2} Assume $\boldsymbol{x} = W \boldsymbol{z} + \boldsymbol{e}$. If $k\mu_W \leq {{|z_{k}|} \over {2|z_1|}} - {{\mu_e} \over {|z_1|}}$, then $\text{supp}(\boldsymbol{z})$ = $\text{supp}(\bar{\boldsymbol{z}})$.

The noisy-case upper bound on $k\mu_W$ turns out to be close to the noiseless bound, if the magnitudes of all nonzero entries in $\boldsymbol{z}$ increase proportionally, so that ${{|z_{k}|} \over {2|z_1|}}$ remains unchanged but ${{\mu_e} \over {|z_1|}}$ vanishes. This is equivalent to improving the signal-to-noise ratio (SNR) of the input signal. 
%
%



\subsection{The Strong Recovery Guarantee}




Recall that the RIP~\cite{candes2005decoding} condition assumes:
\begin{assumption}\label{asm}
For $\forall \boldsymbol{z} \in \Omega_k$, there exists $\delta_W \in (0,1)$ s.t. $(1-\delta_W) \leq \frac{||Wz||^2}{||z||^2} \leq (1+\delta_W)$.



\end{assumption}
We further introduce the stronger guarantee form:
\begin{theorem}\label{thm2}
Assume $\boldsymbol{x} = W \boldsymbol{z} + \boldsymbol{e}$, $\bar{\boldsymbol{z}} = \text{max}_{k}(W^\top \boldsymbol{x})$, $\boldsymbol{z} \in \Omega_k$. If $\delta_W \in (0, 1 - \frac{\sqrt{3}}{2})$, then $\text{supp}(\boldsymbol{z})$ = $\text{supp}(\bar{\boldsymbol{z}})$, given that the smallest nonzero element in $\boldsymbol{z}$ is large enough:
\begin{equation}
 \begin{aligned}
|z_{k}| \ge 2\sqrt{2 \delta_W -  \delta_W^2} ||\boldsymbol{z}||_2 + 2 \left\| \text{max}_{\text{\emph{2k}}}(W^\top \boldsymbol{e}) \right\|. 
  \end{aligned}
  \label{3}
\end{equation}
\end{theorem}

When condition (\ref{3}) holds, a lower bound for $||\boldsymbol{z}||$ can also be derived as:
\begin{equation}
 \begin{aligned}
||\boldsymbol{z}|| \ge \frac{2 \sqrt{k}}{1 - 2\sqrt{k(2 \delta_W -  \delta_W^2)}} \left\| \text{max}_{\text{\emph{2k}}}(W^\top \boldsymbol{e}) \right\|. 
  \end{aligned}
    \label{4}
\end{equation}
To ensure $1 - 2\sqrt{k(2 \delta_W -  \delta_W^2)} > 0$,  $k$ must be less than $\frac{1}{4(2\delta_W-\delta_W^2)}$. Consistent with the weak guarantee case, (\ref{3}) and (\ref{4}) also encourage small $\delta_W$ and $k$, uncorrelated small noise and high SNR. Different from Theorem \ref{thm1}, Theorem \ref{thm2} enforces no extra requirement on the absolute magnitudes of the nonzero entries in $\boldsymbol{z}$ except for a lower bound. It is also noted that Theorem \ref{thm2} is stricter than original RIP~\cite{candes2005decoding} constraint.

\section{DLTF: A Dictionary Learning Model for Thresholded Features}

\subsection{Model Formulation}
As illustrated by Theorems \ref{thm1} and \ref{thm2}, in order to achieve perfect support recovery in the simple thresholded feature, two crucial points are (at least) required in addition to the sparsity of $\boldsymbol{z}$: 1) $W$ has small $\mu_W$ and/or $\delta_W$; 2) the residual $\boldsymbol{e}$ is small and nearly uncorrelated with $W$. Taking them into account, we design a \textit{Dictionary Learning model for Thresholded Features} (DLTF). Specifically, 
we follow \cite{duarte2009learning} to encourage the Gram matrix of $W$ to be close to the identity by minimizing $||W^\top W - I||^2$, which enforces $W$ to have small $\mu_W$ or $\delta_W$. Moreover, Theorem \ref{thm2} suggests to minimize $\left\| \text{max}_{\text{\emph{2k}}}(W^\top \boldsymbol{e}) \right\|$. It is noted that $||\text{max}_{\text{\emph{k}}}(.)||$ is a convex, sub-differentiable vector norm, under the name of $(k,2)$ symmetric gauge norm \cite{bhatia1997matrix}, or $(k,2)$ norm for short \cite{tono2017efficient}. Thus we re-write $||\text{max}_{\text{\emph{2k}}}(W^\top \boldsymbol{e})||^2$ as $||W^\top \boldsymbol{e}||_{2k,2}^2$ in what follows.

Let $X \in \R^{n \times N}$ = $\{\boldsymbol{x}_i\}$ be the training set, and $Z \in \R^{m \times N}$ = $\{\boldsymbol{z}_i\}$ be the corresponding sparse codes. The proposed DLTF approach is to learn $W$ with the following properties: (1) $X$ can be well approximated by $W Z$; (2) $\forall i$, $\text{supp}(\boldsymbol{z}_i)$ and $\text{supp}(\text{max}_{k}(W^\top \boldsymbol{x}_i))$ are as close as possible, which is achieved through Theorems \ref{thm1}, \ref{thm2} by minimizing $||W^\top W - I||^2$ and $||W^\top \boldsymbol{e}||_{2k,2}^2$. In order to achieve the above goal, we formulate the objective function as follows:  

\begin{equation}
 \begin{aligned}
\min_{W, Q, Z} \, \frac{\lambda}{2} \sum_{i=1}^N ||\boldsymbol{q}_i||_{2k,2}^2 & + ||W^\top W - I||^2 + \frac{\theta}{2}\|X- W Z\|^2 \\ s.t.\, & \, Q = W^\top (X- W Z); \\ 
& \, ||\boldsymbol{z}_i ||_0 \le k, , i = 1, 2, ..., N; \\
& ||\boldsymbol{w}_j|| = 1, j = 1,...,m.
  \end{aligned}
 \label{dl}
\end{equation}
where $Q = W^\top (X- W Z) \in \R^{m \times N}$ and $\boldsymbol{q}_i ($i = 1, ..., N$)$ is the $i$-th column. $\lambda, \theta$ are two scalars. We introduce $Q$ to reduce the complexity of $W$ since $W$ is involved in both the non-smooth $(2k,2)$ norm term and the diagonal penalty term simultaneously.


\subsection{Algorithm Development}

\begin{algorithm*}[t]
\caption{Algorithm to solve proximal mapping (\ref{Qproximal}) for ordered and positive vector.}
\label{alg:prox}
\SetAlgoLined
\KwIn{Vector $\boldsymbol{c} \in \R^m, \boldsymbol{c} \geq \0$ and $\boldsymbol{c}$ is in increasing order, parameter $\gamma \geq 0$.}
 \KwResult{Problem solution $\boldsymbol{p}^*$.}
 $\u_{1:m-k}=\boldsymbol{c}_{1:m-k'}, \u_{m-k'+1:m}={1\over 1+\gamma}\boldsymbol{c}_{m-k'+1:m}$;\\
 $\boldsymbol{t}_{1:m-k'}=1, \boldsymbol{t}_{m-k'+1:m}=1+\gamma$;\\
 $\boldsymbol{p}^*=$ Reduce($\boldsymbol{u},\boldsymbol{t},1$);\\
  \algrule[1pt]
\SetKwFunction{proxfn}{\text{Reduce}}
\Indm\nonl\proxfn{$\boldsymbol{u},\boldsymbol{t},j$}\\
\Indp
Let $J$ be the dimension of $\u$;\\
 \While{$j \leq J$}
 {
 \If{$\u_{j} > \u_{j+1}$}
 {
   $\boldsymbol{u}'=[\boldsymbol{u}_{1:j-1}, {\boldsymbol{t}_j\boldsymbol{u}_j + \boldsymbol{t}_{j+1}\boldsymbol{u}_{j+1} \over \boldsymbol{t}_j +\boldsymbol{t}_{j+1}} ,\boldsymbol{u}_{j+2:end}]$ \tcp*{Remove $\u_{j+1}$}
   $\boldsymbol{t}'=[\boldsymbol{t}_{1:j-1}, \boldsymbol{t}_j + \boldsymbol{t}_{j+1}, \boldsymbol{t}_{j+2:end}]$\tcp*{Remove $\boldsymbol{t}_{j+1}$}
   $\boldsymbol{x} =$ Reduce($\boldsymbol{u}',\boldsymbol{t}',\text{max}(1, j-1)$) \tcp*{Recursively invoke Reduce}
   \KwRet{$[\x_{1:j}, \x_j, \x_{j+1:end}]$} \tcp*{Duplicate $\x_{j}$ since $\x_{j}=\x_{j+1}$}
 }
 $j=j+1$;\\
 }
 \KwRet{\u}
\end{algorithm*}

We apply the optimization framework of Alternating Direction Method of Multipliers (ADMM). The augmented Lagrangian function of (\ref{dl}) is:

\begin{equation}
\begin{aligned}\label{lag}
& \quad \quad \frac{\lambda}{2} \sum_{i=1}^N ||\boldsymbol{q}_i||_{2k,2}^2 + ||W^\top W - I||^2 + \frac{\theta}{2}\|X- W Z\|^2 \\
& + \langle Y, Q-W^\top (X- W Z) \rangle + \frac{\beta}{2}||Q-W^\top (X- W Z)||^2 \\
& \hspace{2cm} s.t.  \, ||\boldsymbol{z}_i ||_0 \le k, , i = 1, 2, ..., N; \\
& \hspace{2.5cm} ||\boldsymbol{w}_j|| = 1, j = 1, 2, ..., m.
\end{aligned}
\end{equation}
where $Y \in \R^{m \times N}$ is the Lagrange multiplier and $\beta$ is a positive constant. We then sequentially solve the three subproblems at the $t$-th iteration ($t$ = 0, 1, ...). 

\subsubsection{$Z$-subproblem} Solving $Z$ is a standard sparse decomposition problem, which can be solved separately for each $\boldsymbol{z}_i$ using the iterative algorithm \cite{blumensath2008iterative}:

\begin{equation}
\begin{aligned}
Z_{t+1} &  = \arg \min_{Z} \, \frac{\theta}{2}\|X- W_t Z\|^2 + \langle Y_t, W_t^\top W_t Z \rangle \\ 
   & + \frac{\beta}{2}||W_t^\top W_t Z - W_t^\top X + Q_t||^2 \\
& s.t.\, ||\boldsymbol{z}_i ||_0 \le k, , i = 1, 2, ..., N.
 \label{subZ}
\end{aligned}
\end{equation}

\subsubsection{$Q$-subproblem}
The $Q$ update could also be solved separately for each $\boldsymbol{q}_i$:

\begin{equation}
 \begin{aligned}
Q_{t+1} & = \arg \min_{Q} \, \sum_{i=1}^N ||\boldsymbol{q}_i||_{2k,2}^2 \\
   & +  \frac{\beta}{\lambda}||Q-(W_t^\top X- W_t^\top W_t Z_{t+1} - \frac{{Y}_t}{\beta})||^2
  \end{aligned}
 \label{subQ}
\end{equation}
Let $\gamma \geq 0$, and $\boldsymbol{q}, \boldsymbol{c} \in \R^m$.  Define the proximal mapping of the $(k',2)$ norm for $\boldsymbol{q}$:

\begin{equation}
 \begin{aligned}
 prox^{k',2}_{\gamma} (\boldsymbol{c}) = \arg\min_{\boldsymbol{q}} \gamma||\boldsymbol{q}||_{k',2}^2 +  ||\boldsymbol{q} - \boldsymbol{c}||^2
  \end{aligned}
 \label{Qproximal}
\end{equation}
Problem~\eqref{subQ} is converted to solving the proximal mapping~\eqref{Qproximal} with $\gamma = \frac{\lambda}{\beta}$ and $k'=2k$. To our best knowledge, only the basic subgradient method \cite{tono2017efficient} was exploited for the optimization of ($k$,2) norm in literature. We present an efficient $O(m\log m)$ solution for~\eqref{subQ} as described in next section.

\subsubsection{$W$-subproblem}
The $W$ update solves the following manifold constrained problem:

\begin{equation}
 \begin{aligned}
 W_{t+1} & = \arg \min_{W} \, ||W^\top W - I||^2 + \frac{\theta}{2}\|X- W Z_{t+1}\|^2 \\
& - \langle Y, W^\top (X- W Z_{t+1}) \rangle  \\
& + \frac{\beta}{2}||Q_{t+1}-W^\top (X- W Z_{t+1})||^2 \\
& \quad s.t.\, ||\boldsymbol{w}_j|| = 1, j = 1, 2, ..., m.
\end{aligned}
\label{subW}
\end{equation}
We apply the curvilinear search algorithm in \cite{wen2013feasible} to solve~(\ref{subW}) as it lies in the spherical constraint.


Furthermore, $Y$ is updated as: ${Y}_{t+1} = {Y}_{t} +  \beta(Q_{t+1}- W_{t+1}^\top X + W_{t+1}^\top W_{t+1} Z_{t+1})$.

\section{Efficient $O(m\log m)$ Proximal Mapping of ($k$, 2) Norm}

It is noted that solving~\eqref{Qproximal} with subgradient descent is yet inefficient. Therefore, we propose an efficient proximal algorithm for the ($k$, 2) norm, which also contributes the paper.
\begin{theorem}
\label{thm:Qproximal}
The proximal mapping~\eqref{Qproximal} is solved by Algorithm~\ref{alg:prox} in $O(m\log m)$ time complexity.
\end{theorem}

\noindent \textbf{Proof sketch:} To prove Theorem~\ref{thm:Qproximal}, we first establish:
\begin{lemma}
\label{lem:order}
For the problem~\eqref{Qproximal} with $\boldsymbol{c} \geq \0$, the order of coordinates in optimal solution $\boldsymbol{q}^*$ is the same as the order of the corresponding coordinates in $\boldsymbol{c}$.
\end{lemma}
Lemma \ref{lem:order} shows that the proximal mapping~\eqref{Qproximal} will not change the sign of $\boldsymbol{c}$, i.e., for all $i$, $\text{sign}(prox^{k',2}_{\gamma} (\boldsymbol{c})_i) = \text{sign}(c_i)$. Then we only need consider the magnitude of entries in $\boldsymbol{c}$. We can sort the entries of $\boldsymbol{c}$ (in magnitude). Therefore, with additional time complexity $O(m\log m)$ for sorting, we can convert~\eqref{Qproximal} with any vector $\boldsymbol{c}\in \R^m$ to the proximal mapping for ordered and positive vector. 

We then introduce the following lemma: 
\begin{lemma}\label{lem:rec}
Optimization problem
\begin{gather}
\min_{{\bf x} \in \R^J} \sum_{j=1}^J {\bf t}_j({\bf x}_j - \u_j)^2 \label{eq:orderobj}\\
\text{s. t. } {\bf x}_1 \leq {\bf x}_2 \leq ... \leq {\bf x}_J \label{eq:ordercons}
\end{gather}
can be solved by invoking the subroutine ``Reduce($\boldsymbol{u}$, $\boldsymbol{t}$, 1)'' in Algorithm~\ref{alg:prox}.
\end{lemma}

To solve \eqref{eq:orderobj}, the key step is applying Lemma~\ref{lem:merge} below to iteratively merge neighbor variables to obtain a reduced problem. When the reduced problem has input $\u'$ containing monotonically increasing elements, the solution $\x'=\u'$.
\begin{lemma}\label{lem:merge}
If $\u_j > \u_{j+1}$, then the optimal solution $\x^*$ to~\eqref{eq:orderobj} should satisfy $\x^*_j = \x^*_{j+1}$.
\end{lemma}

We describe the detailed proof of Lemmas \ref{lem:order}, \ref{lem:rec}, and \ref{lem:merge} in the appendix. 

\section{Experiments}

The main purpose of this section is to demonstrate that DLTF possesses the capability to learn the dictionary, that reliably recovers the sparse support (see synthetic experiments) and benefits the practical utilization of thresholded features most (see real data experiments). It is important to note that DLTF is not ``yet another'' way of \textit{standard dictionary learning}, whose inference relies on iterative sparse solvers. Instead, it is a new type of dictionary learning, specifically designed for using the thresholded feature (\ref{feature}) at inference, which prioritize (extreme) \underline{simplicity}, \underline{efficiency}, \underline{general usability} and \underline{theoretical guarantees}. Despite not being optimized for any specific task, DLTF shows competitive performance for a variety of real-data tasks with minimal time costs. 

We strive to compare DLTF with standard dictionary learning algorithms, and choose the popular KSVD \cite{KSVD} as a representative of the latter in most experiments. In a few real-data experiments (e.g, clustering), we compare the dictionaries computed by DLTF and KSVD, and use them to compute both thresholded features, and canonical sparse codes (via iterative algorithms). \textbf{To avoid confusion}, such comparison is intended as an ``ablation study'' (altering training and testing components) to show that: (1) DLTF dictionary is much better suited for the thresholded feature than conventional dictionaries; (2) the results of applying thresholded feature solved with DLTF dictionary are superior to or comparable with applying sparse features solved with more expensive iterative algorithms. It does not contradict our default pipeline of training DLTF dictionary and computing thresholded features at inference.

\begin{table*}[!htp]
\caption{The support recovery performance comparison, at different sparsity levels $k$, measured by \textit{ave\_dif}. The DLTF results are reported with $\lambda = 0.05$, $\theta = 0.01$.}
\centering{
\setlength{\tabcolsep}{15pt}
\resizebox{0.75\textwidth}{!}{
\begin{tabular}{| c | c | c | c | c | c |}
\hline
 & $k = 4$ & $k = 6$ & $k = 8$ & $k = 10$ & $k = 12$  \\
\hline
original   & 0.354 & 0.778 & 1.241 & 1.746 &  2.293 \\
\hline
random   & 3.876 & 5.721 & 7.499 & 9.204 & 10.895  \\
\hline
KSVD   & 0.724 & 1.988 & 3.897 & 6.110 & 7.430  \\
\hline
DLTF       & 0.495 & 1.119 & 1.879 & 2.753 & 3.759 \\
\hline
\end{tabular}}
}
\label{tab::sym_ave_dif}
\end{table*}

\subsection{Support Recovery in Thresholded Features: Synthetic Simulations}



We first evaluate the performance of the support recovery on synthetic data. We generate an over-complete i.i.d. random Gaussian matrix as the dictionary $W_0 \in \R^{n\times m}$, and the sparse codes $Z \in \R^{m \times N}$ = $\{\boldsymbol{z}_i\}$, where each $\boldsymbol{z}_i$ has only $k$ nonzero entries with the value 1 and random locations. We then synthesize $X \in \R^{n \times N}$ = $\{\boldsymbol{x}_i\}$ by: $X = W_0 Z + E$, where each entry of the noise matrix $E$ is i.i.d. sampled from $\mathcal{N}(0, 0.01)$. By default, we fix $n = 64$, $m = 128$, and $N$ = 10, 000 for the training set. A testing set of 10, 000 samples are generated separately.  

In order to compute the thresholded features $\bar{\boldsymbol{z}_i} = \text{max}_{k}(W^\top \boldsymbol{x}_i)$, we compare DLTF with three baselines: two are intentionally chosen to roughly indicate the empirical performance ``upper bound'' and ``lower bound'', as well as directly employing the conventional KSVD dictionary:
\begin{itemize}
\item  \textit{Original} baseline: $W$ = $W_0$. Note that the random Gaussian $W_0$ itself has very small mutual incoherence/RIP constant, and it is known that $X$ can be sparsely represented over $W_0$. Therefore, we expect $W_0$ to be nearly an optimal solution to (\ref{dl}), and this original baseline's results will be likely close to the best attainable support recovery performance. 
\item \textit{Random} baseline: $W$ = $W_r$ which is another independent random Gaussian matrix.
\item \textit{KSVD} baseline: $W$ = $W_K$ is learned from $X$ by KSVD and then applied to thresholded features. 
\end{itemize}

For DLTF, we denote $W = W_{\text{DLTF}}$ as solved from (\ref{dl}), using random initializations. It is noted that the original baseline is an ``ideal'' case that exists only in simulations. Except for a handful of cases such as compressive sensing \cite{baraniuk2007compressive}, it is unlikely to have such a pre-known dictionary in practice, over which the target signals can be accurately represented, and whose mutual coherence/RIP constant is as small as the random bases. 

\begin{figure*}
\centering{
\begin{subfigure}{0.33\textwidth}
  \centering
  \includegraphics[width = 2in,height = 1.55in]{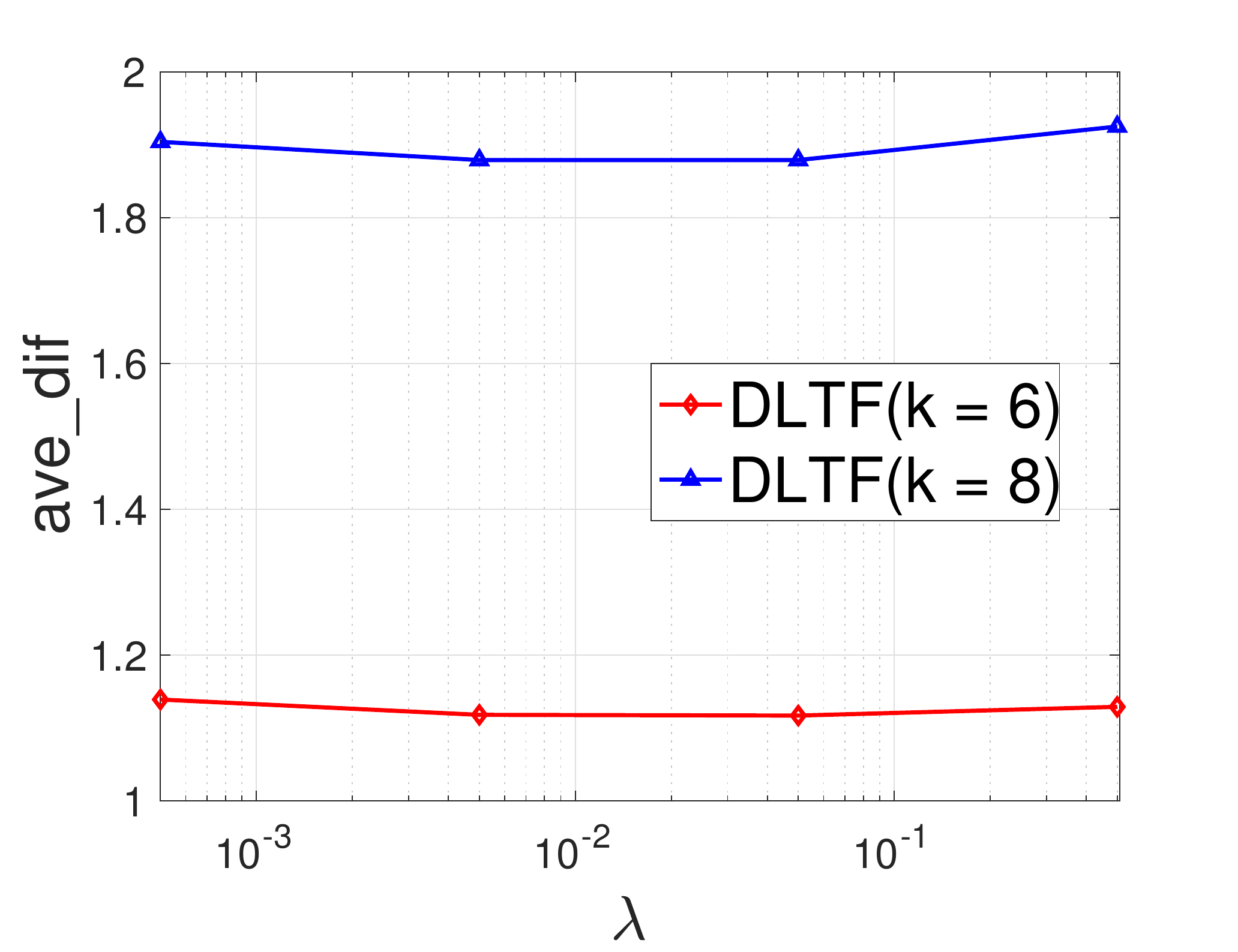}
  \caption{}
  \label{fig:lambda}
\end{subfigure}%
\begin{subfigure}{0.33\textwidth}
  \centering
  \includegraphics[width = 2in,height = 1.55in]{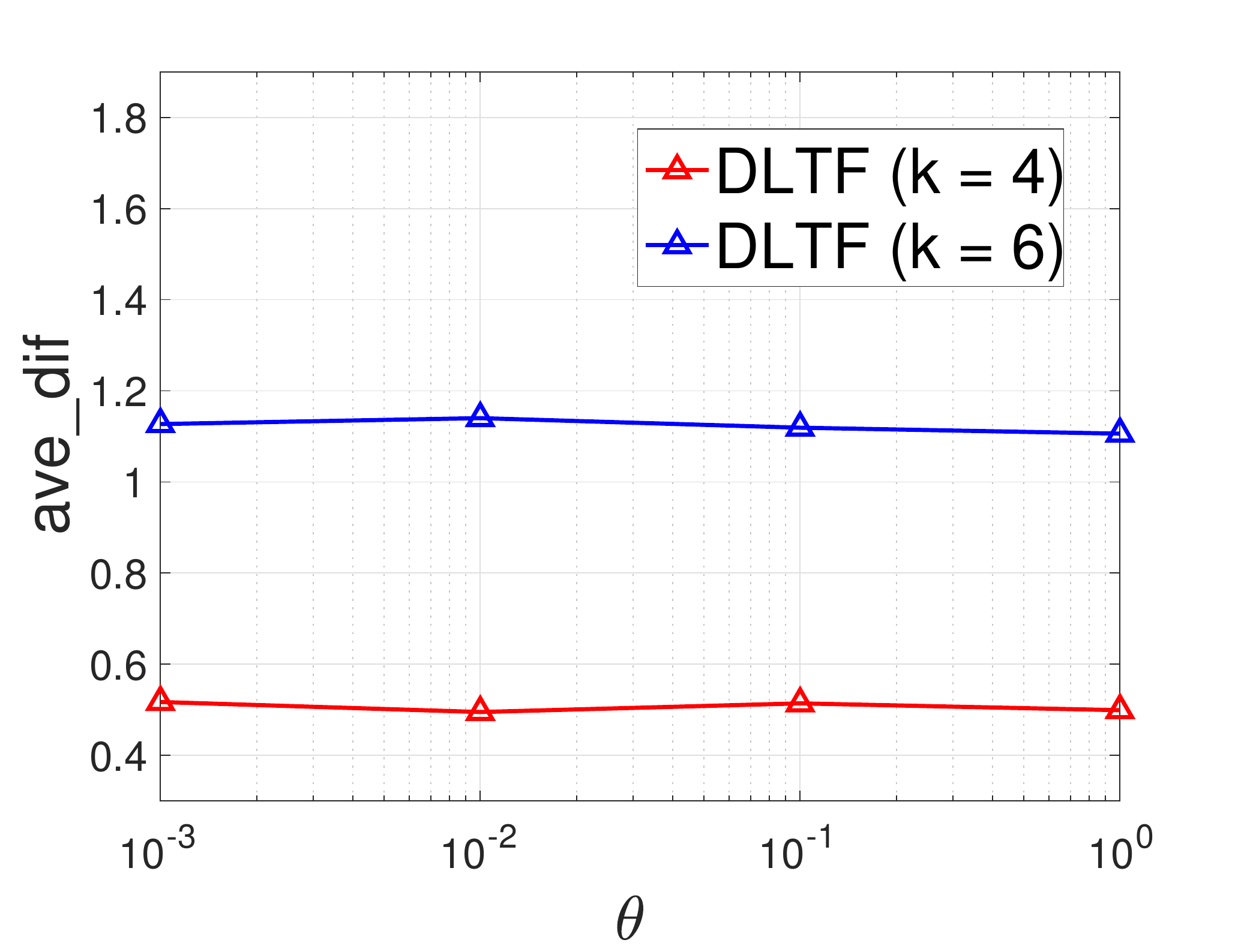}
  \caption{}
  \label{fig:theta}
\end{subfigure}%
\begin{subfigure}{0.33\textwidth}
  \centering
  \includegraphics[width = 2in,height = 1.55in]{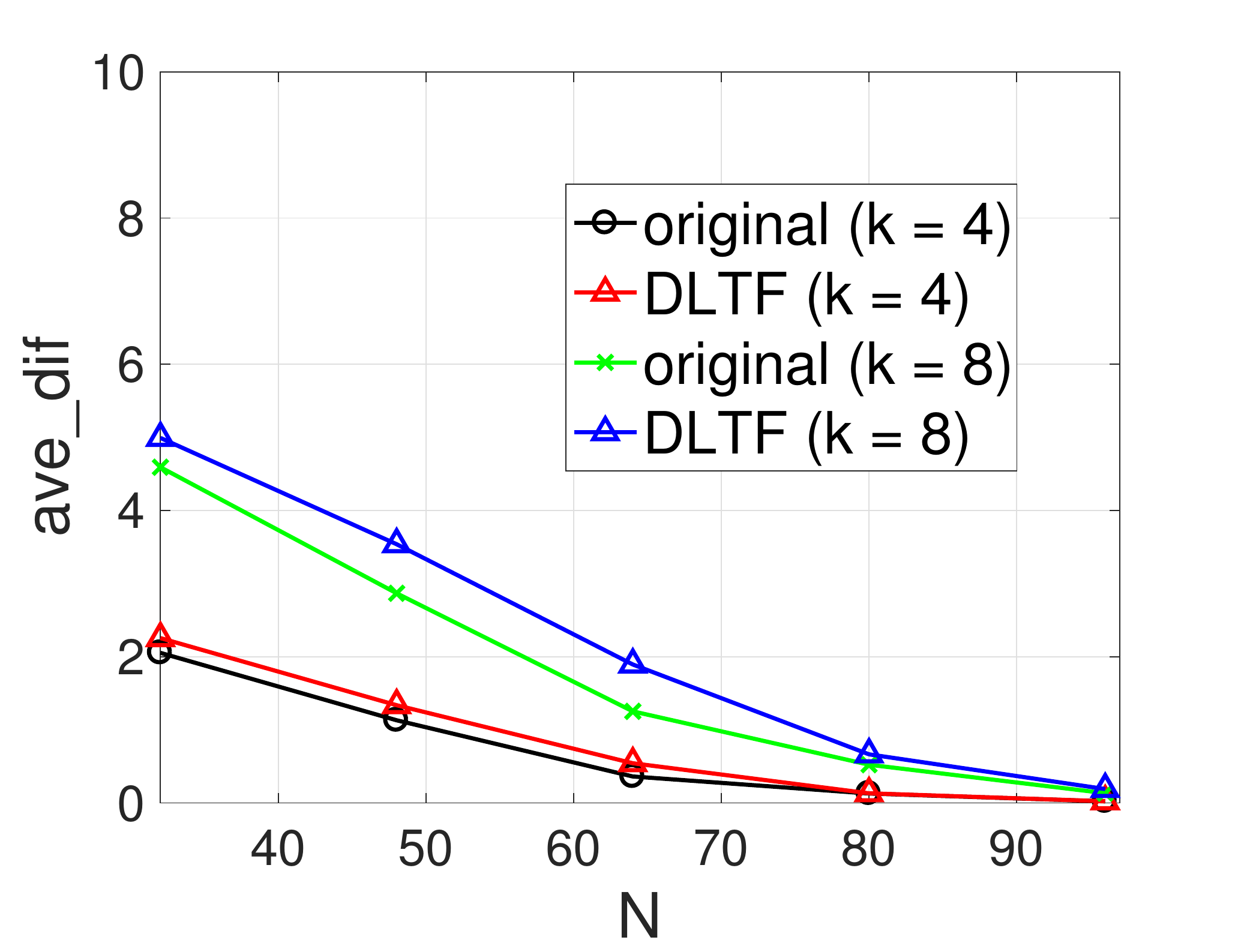}
  \caption{}
  \label{fig:feature_dim}
\end{subfigure}%
\newline
\begin{subfigure}{0.33\textwidth}
  \centering
  \includegraphics[width = 2in,height = 1.55in]{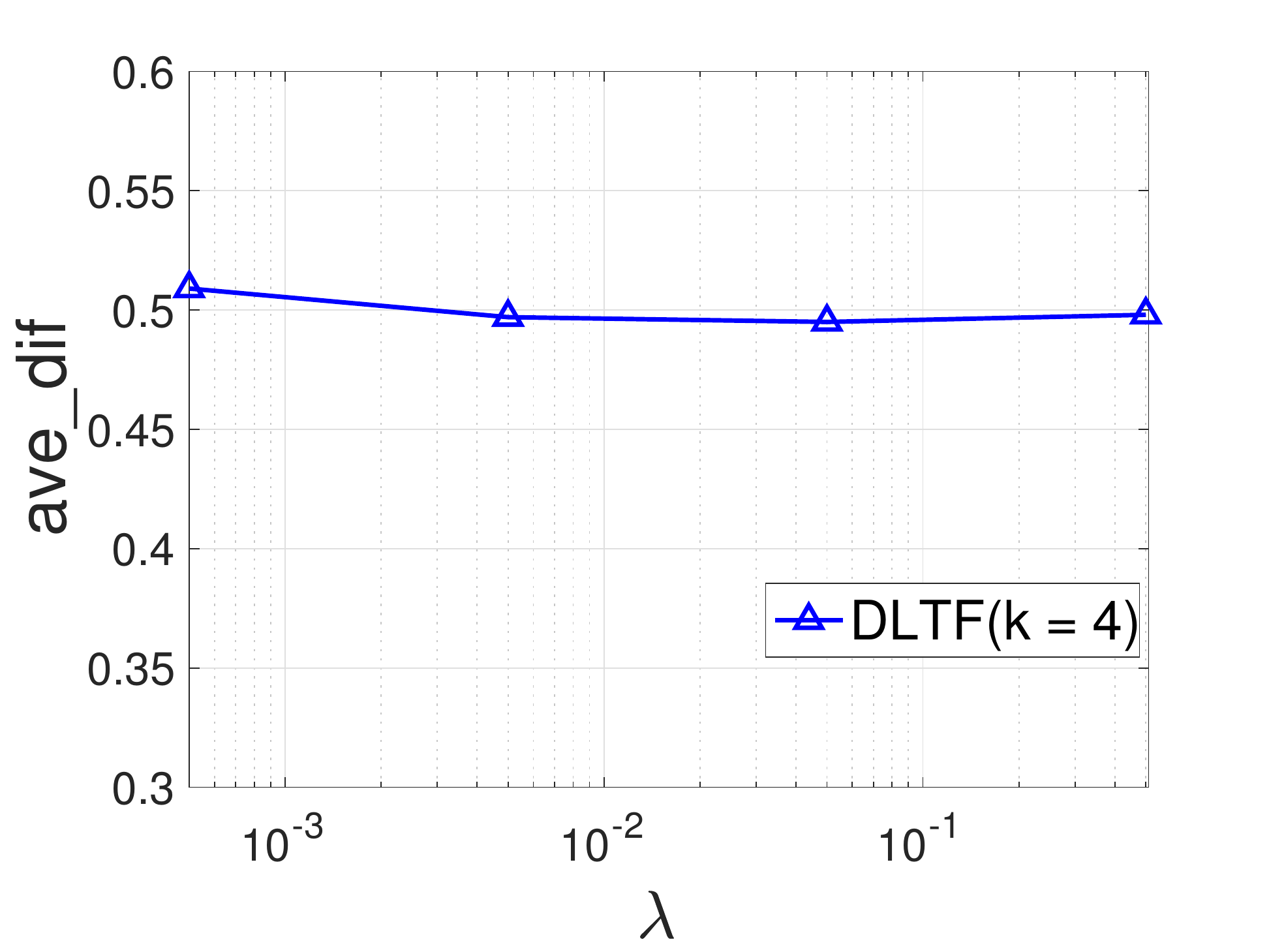}
  \caption{}
  \label{fig:lambda_4}
\end{subfigure}%
\begin{subfigure}{0.33\textwidth}
  \centering
  \includegraphics[width = 2in,height = 1.55in]{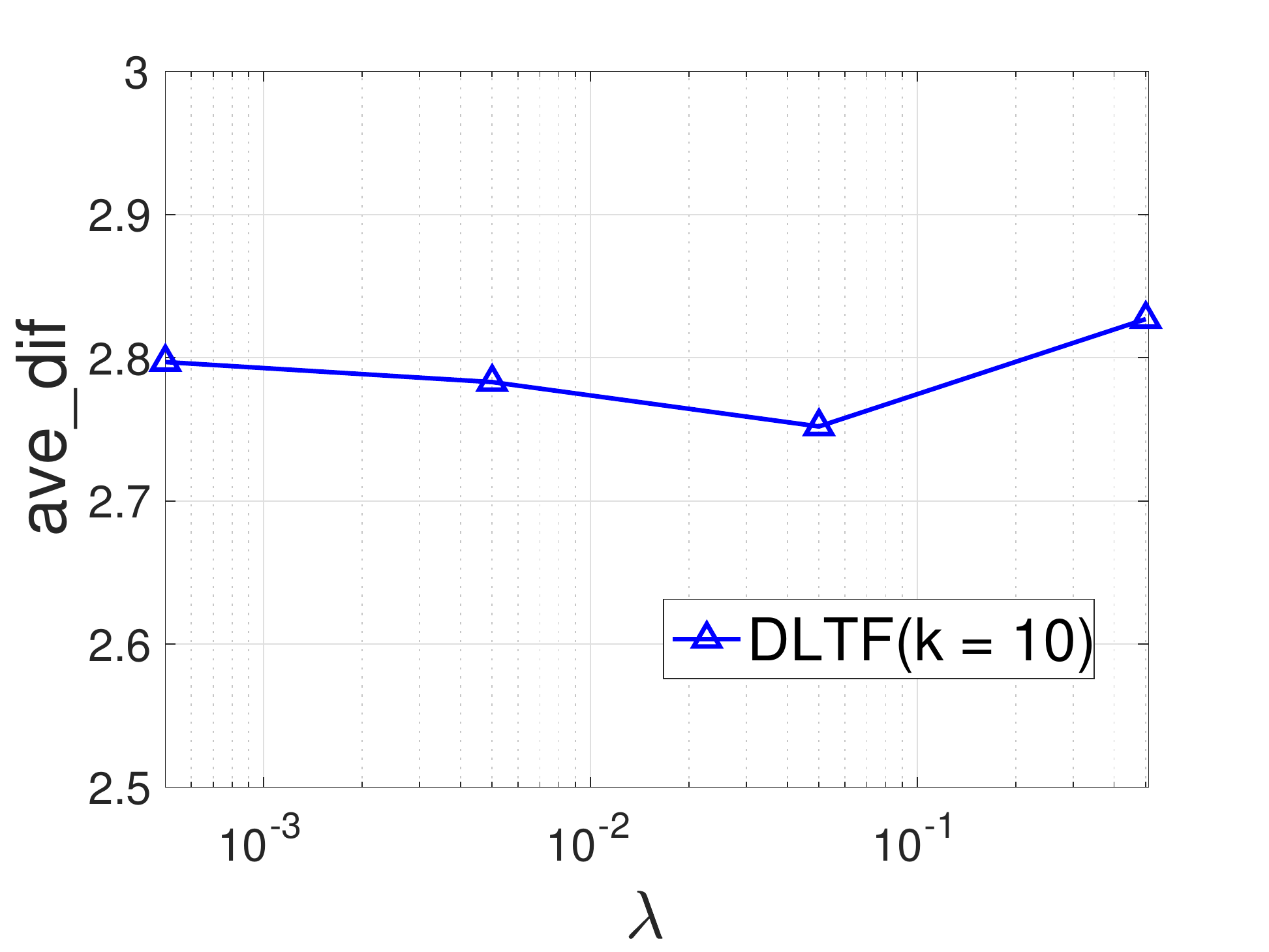}
  \caption{}
  \label{fig:lambda_10}
\end{subfigure}%
\begin{subfigure}{0.33\textwidth}
  \centering
  \includegraphics[width = 2in,height = 1.55in]{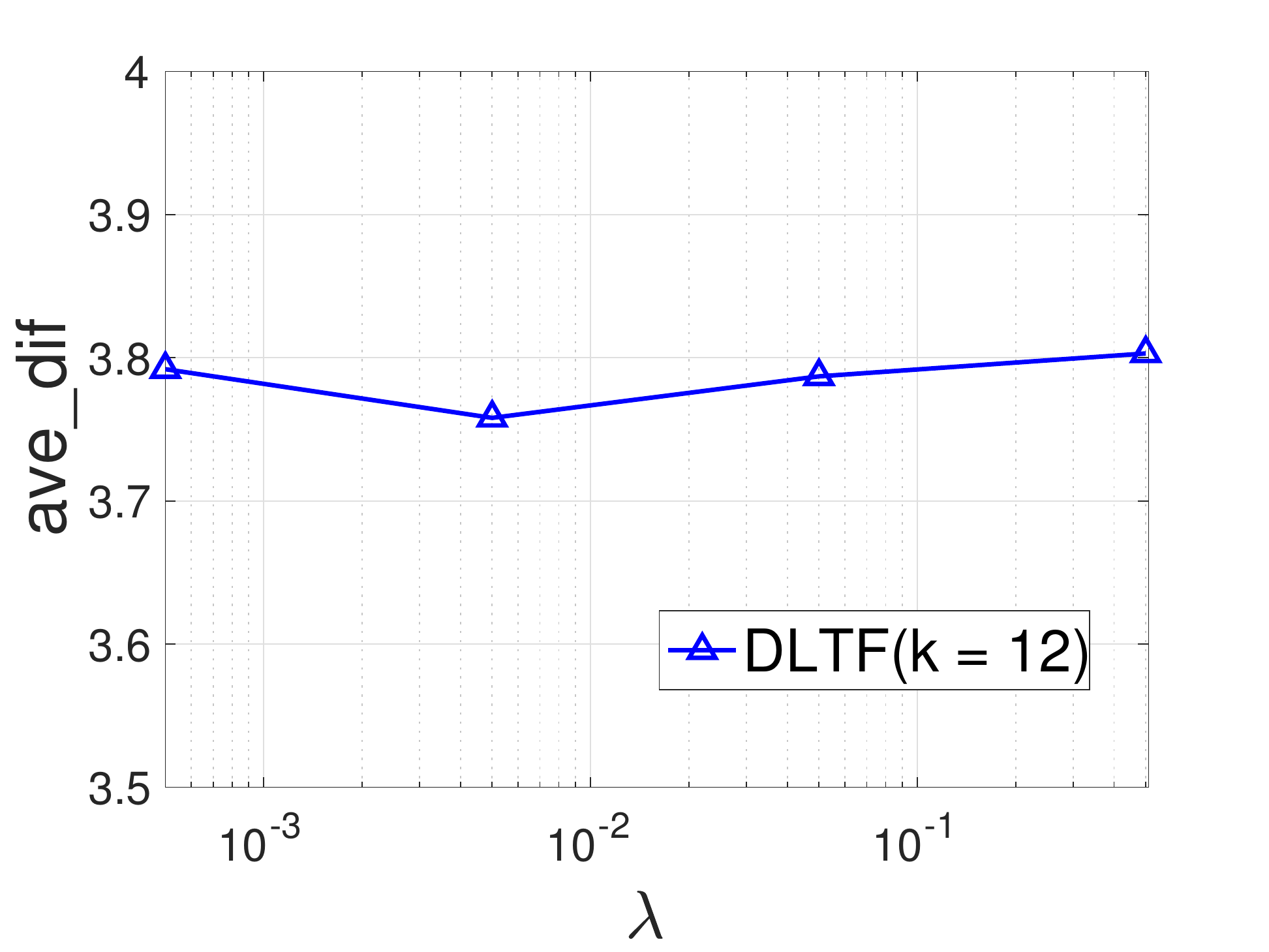}
  \caption{}
  \label{fig:lambda_12}
\end{subfigure}%
}
\caption{The support recovery performance comparison by varying:  (a) the DLTF hyper-parameter $\lambda$; (b) the DLTF hyper-parameter $\theta$; (c) the observed data dimension $n$. More comparison by varying $\lambda$ for the following values: (d) $k$ = 4; (e) $k$ = 10; (f) $k$ = 12.}
\label{fig::parameter_sensitivity}
\end{figure*}

To evaluate the accuracy of the recovered support, we define the following metric that calculates the averaged per-sample support differences between $\bar{\boldsymbol{z}_i}$ and $\boldsymbol{z}_i$: 
\begin{equation}
 \begin{aligned}
\textit{ave\_dif} = \frac{1}{N} \sum_{i = 1} ^ {N} \frac{|\text{supp}(\bar{\boldsymbol{z}}_i) \oplus \text{supp},(\boldsymbol{z}_i)|_1}{2}
\label{ave_dif}
  \end{aligned}
\end{equation}
where $\oplus$ denotes the element-wise XOR operation. The value of \textit{ave\_dif} ranges between $[0, k]$.

We first vary the sparsity level $k \in  [4, 6, 8, 10, 12]$, and compare the support recovery results in Table~\ref{tab::sym_ave_dif}. It can be seen that DLTF achieves overall comparable results to the original baseline, especially when $k$ is small. The performance gap slightly increases as $k$ grows up. Compared to the original baseline with knowing $W_0$ as a prior, DLTF achieves competitive performance by only observing $X$. KSVD dictionaries perform poorly when applied to computing thresholded features, although the random baseline performs the worst. The above result clearly supports the necessity of DLTF. 

\subsection{Ablation Study}
Next, we investigate the effects of varying several (hyper-)parameters in Figure~\ref{fig::parameter_sensitivity}(a)(b), all of which further verify the effectiveness and robustness of DLTF. It can be seen that DLTF maintains stable performance over a wide range of ($\lambda, \theta$) values. When $\lambda$ turns either too small or too large, the DLTF performance will be degraded a bit, which manifests the trade-off between the regularization effects of the first term (uncorrelated small noise) and the second term (small mutual coherence/RIP constant of $W$) in (\ref{dl}). Moreover, we further present the support recovery performance comparison plots by varying $\lambda$, at $k$ = 4, 10 and 12, respectively in Figure~\ref{fig::parameter_sensitivity} (d)(e)(f). The observations and conclusions are similar to the $k$ = 6 and $k$ = 8 plots in Figure~\ref{fig::parameter_sensitivity} (a)(b). We omit the same plots for $\theta$, as we find the DLTF performance to be insensitive to varying $\theta$ at all $k$ values.  We choose $\lambda$ = 0.5, $\theta$ = 0.01 as default values hereinafter. 

It is also noted that Figure~\ref{fig::parameter_sensitivity}(c) varies $n \in  [32, 48, 64, 80, 108]$, when $m$ = 128 is fixed. Both the original baseline and DLTF benefit from increasing $n/m$ ratio, and their performance difference seems to vanish as $n/m \rightarrow 1$.

Finally, we compute the mutual coherence of $W_0$ and $W_{\text{DLTF}}$, at $m$ = 128, $n$ = 64, which are approximately 0.49 and 0.56, respectively\footnote{Results possess certain randomness. We did not compute the RIP constant as its calculation is NP-hard.}, respectively. It indicates that DLTF indeed finds a solution $W$ with low mutual coherence. Besides, Corollary 1.1 suggests a sufficient recovery condition of $\mu_W \le \frac{1}{2k}$, which cannot be met in this case for $\forall k > 1$. While the condition is not necessary, it implies that our simulation settings are challenging.

\subsubsection{Complexity and Running Time Analysis}
The time complexity of computing the thresholded feature by (\ref{feature}) for $N$ samples is $\mathcal{O}(N (mn+m+k\log m))$, which is roughly equal to the complexity of running iterative sparse solvers e.g., orthogonal matching pursuit (OMP) as used in \cite{KSVD}, \underline{for only one iteration}. In practice, implementation differences, hyper-parameter choices and stop conditions\footnote{Iterative algorithms may be stopped by measuring residual fitting errors, maximum sparsity, or iteration numbers, etc.} can dramatically affect the efficiency of iterative sparse solvers. However, it is self-evident that the efficiency advantage of (\ref{feature}) over iterative solvers is independent of implementations. 

\begin{table*}[htp!]
\caption{The ACC and NMI comparison between DLTF and KSVD methods w.r.t. different $k$ values, and the total testing time (in seconds) comparison, of different approaches.}
\centering{
\setlength{\tabcolsep}{15pt}
\resizebox{0.75\textwidth}{!}{
\begin{tabular}{| c | c | c | c | c | c|}
\hline
\multicolumn{2}{|c|}{Training}  & DLTF & DLTF & KSVD & KSVD \\
\hline
\multicolumn{2}{|c|}{Testing} & Exact & TF & Exact & TF \\
\hline
\hline
\multirow{2}{*}{$k$ = 5} & ACC & 0.532 & 0.545 & 0.524 &  0.517  \\
\cline{2-6}
 & NMI & 0.525 & 0.532  &  0.526 & 0.521 \\
\hline
\multirow{2}{*}{$k$ = 10} & ACC & 0.584 & 0.557 & 0.529 &  0.524  \\
\cline{2-6}
 & NMI  &  0.552 & 0.525  &  0.520 & 0.517  \\
 \hline
 \multirow{2}{*}{$k$ = 30} & ACC  & 0.585 & \textbf{0.594} & 0.596 &  0.586   \\
\cline{2-6}
 & NMI  &  0.545 & \textbf{0.550}  &  0.561 & 0.533 \\
 \hline
 \multirow{2}{*}{$k$ = 40} & ACC  & 0.572 & 0.567 & 0.599 &  0.591  \\
\cline{2-6}
 & NMI  &  0.534 & 0.531  &  0.585 & 0.536 \\
\hline
\hline
\multicolumn{2}{|c|}{Testing Time (s)}  & 3.639 & \textbf{0.394} & 2.851 & 0.409 \\ 
\hline
\end{tabular}}}
\label{clustering}
\end{table*}

\begin{table*}[h]
\caption{The ACC, NMI and testing time comparison between DLTF TF ($k$ = 30) and several other approaches.}
\centering{
\setlength{\tabcolsep}{15pt}
\resizebox{0.75\textwidth}{!}{
\begin{tabular}{| c | c | c | c | c |c|}
\hline
Method  & DLTF TF & KM &  AE-1 & AE-2 & DEC\\
\hline
\hline
ACC & 0.594 & 0.484 & 0.507 & 0.571 & 0.762 \\
\hline
NMI & 0.550 & 0.483 & 0.501 & 0.531 & 0.738\\
\hline
\hline
Time (s) & 0.387 & 0.241 & 0.414 & 0.767 & 1.727 \\ 
\hline
\end{tabular}}}
\label{clustering2}
\end{table*}

In all our experiments, we generally obverse the running time of thresholded features to be one or even two orders of magnitudes less than iterative comparison methods, e.g., KSVD (using OMP for testing). Our default testing environment is Matlab 2016b on a Macbook Pro with 2.7GHz Intel Core i5 CPUs. For example, in the synthetic experiment of $k$ = 8, computing thresholded feature for $N$ = 10,000 samples only costs \textbf{0.25s}. In comparison, solving OMP for the same $N$ = 10,000 samples takes \textbf{3.69s}.

\subsection{Experiments on Image Clustering}

Clustering is an unsupervised task for which sparse codes are known to be effective features \cite{zheng2011graph}. We conduct our clustering experiments on a publicly available subset of MNIST\footnote{\url{http://www.cad.zju.edu.cn/home/dengcai/Data/MLData.html}},  where the first 10, 000 training images of the original MNIST benchmark constitute the training set. A separate set of 10, 000 images is used as the testing set to evaluate the generalization performance. We reshape each 28 $\times$ 28 image into a vector, constructing $X \in \R^{n \times N}$ where $n$ = 784, $N$ =  10, 000. 

We choose $m = 400$, and first design four methods for comparison, by altering ways of training (i.e., learning dictionary via DLTF or KSVD) and testing (i.e., computing sparse codes iteratively or thresholded features):
\begin{itemize}

\item \textit{DLTF Exact:} we apply DLTF to learn $W_{\text{DLTF}} \in \R^{n \times m}$ from the training set $X$, and then exactly solve sparse code features of testing images w.r.t. $W_{\text{DLTF}}$ to global optima, using the iterative algorithm \cite{blumensath2008iterative}. K-Means clustering is applied to cluster the sparse codes features.

\item \textit{DLTF TF:} we apply DLTF to learn $W_{\text{DLTF}}$ from the training set $X$, and then compute the threshold features of the testing images via (\ref{feature}), w.r.t. $W_{\text{DLTF}}$. K-Means clustering is applied to cluster the thresholded features.

\item \textit{KSVD Exact:} replacing DLTF with KSVD \cite{KSVD} in the DLTF Exact approach.

\item \textit{KSVD TF:} replacing DLTF with KSVD in DLTF TF. 

\end{itemize}
In order to validate the effectiveness and efficiency of thresholded features and DLTF, two ``exact'' baselines are further compared. Two standard clustering metrics, the accuracy (ACC) and normalized mutual information(NMI) are used~\cite{zheng2011graph}. Larger ACC and NMI indicate better clustering performance. 


We vary $k \in  [5, 10, 30, 40]$ and report results in Table \ref{clustering}. We first compare two ``TF'' methods. DLTF TF is observed to outperform KSVD TF at most $k$ values, and achieves the best ACC/NMI \underline{among all TF results} at $k = 30$. Second, taking two ``Exact'' baselines into account, it is encouraging to see that DLTF TF consistently achieves similar performance to both, sometimes even superior. Interestingly, we find DLTF TF to usually perform comparably to, or even better than DLTF Exact ($k = 10, 30$). In contrast, KSVD TF always obtains inferior performance than KSVD Exact. The observation demonstrates that $W_{\text{DLTF}}$ is better suited for thresholded features. 

DLTF TF achieves its best performance at $k$ = 30. In comparison, the two KSVD methods seem to favor larger $k$ = 40. To understand why DLTF TF may prefer smaller $k$, the reason may be that the support recovery is more reliable with higher sparsity, based on Theorems \ref{thm1}, \ref{thm2}. On the other hand, small $k$ may cause information loss in sparse codes. A ``medium'' $k$ of 30 seems to best balance the trade-off here.

Moreover, the efficiency of DLTF/thresholded features is further evidenced by the total running time (averaged over different $k$ cases) 
on the 10, 000-sample testing set. The running times of the thresholded feature (both DLTF TF and KSVD TF) are one order of magnitude faster than their exact counterparts. DLTF hence possesses the most competitive performance-efficiency trade-off among the four.



\begin{figure*}[htbp]
\centering
\begin{minipage}{0.32\textwidth}
\centering{
\includegraphics[width=\textwidth]{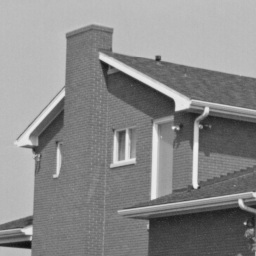}
}\end{minipage}
\begin{minipage}{0.32\textwidth}
\centering{
\includegraphics[width=\textwidth]{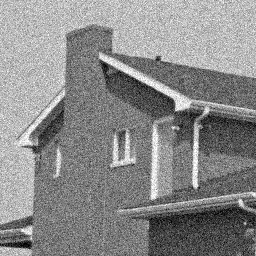}
}\end{minipage}
\begin{minipage}{0.32\textwidth}
\centering{
\includegraphics[width=\textwidth]{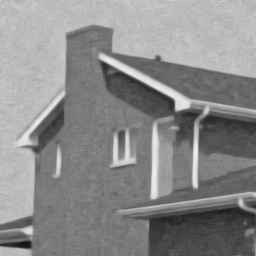}
}\end{minipage}
\\
\begin{minipage}{0.32\textwidth}
\centering{
\includegraphics[width=\textwidth]{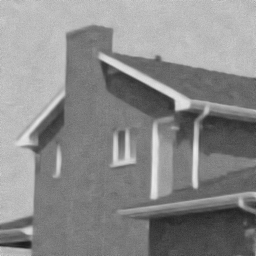}
}\end{minipage}
\begin{minipage}{0.32\textwidth}
\centering{
\includegraphics[width=\textwidth]{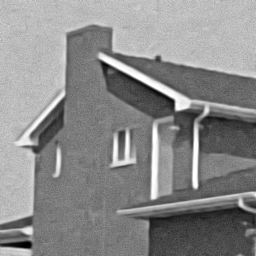}
}\end{minipage}
\begin{minipage}{0.32\textwidth}
\centering{
\includegraphics[width=\textwidth]{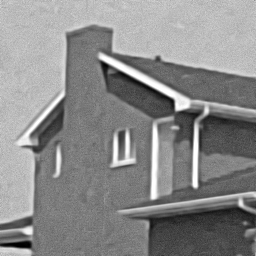}
}\end{minipage}
\caption{Visual comparison on the \textit{House} image, from top left to bottom right: the original image; the noisy image with $\sigma$ =20 (PSNR = 22.09dB); the denoising result using KSVD (PSNR = 31.44dB); the denoising results using DLTF, at $k$ = 1 (PSNR = 30.04 dB), $k$ = 2 (PSNR = 28.56 dB), and $k$ = 4 (PSNR = 27.97 dB).}

\label{noise}
\end{figure*}

We next compare DLTF TF (with the best $k=30$) with other types of clustering models, from the simplest/fastest K-Means to sophisticated neural network (NN) clustering models that typically involve very high complexity:
\begin{itemize}

\item \textit{KM:} we first obtain a PCA matrix $\in \R^{m \times n}$ from $X$, and apply it to reduce the dimension of each testing image to $\R^m$. K-Means clustering is then applied.

\item \textit{AE-1:} an auto-encoder (AE) with one hidden layer $\in \R^m$. ReLU is used. K-Means clustering is applied to the hidden activations of the testing images. 

\item \textit{AE-2:} an AE with two hidden layers, both $\in R^m$. K-Means clustering is applied to cluster the activations of the second hidden layer.

\item \textit{Deep Embedded Clustering (DEC):} a latest deep network clustering model that simultaneously learns feature representations and cluster assignments \cite{DEC}. We used their original MNIST model structure: 784–500–500–2000–10. Note that DEC is much more heavily parametrized than DLTF and other baselines\footnote{DEC was originally trained and evaluated on the full MNIST set of 70,000 samples. For fair comparison with others, we re-train DEC on the given 10,000-sample training set and test on the 10,000-sample testing set.}.

\end{itemize}
To ensure a fair comparison with non-NN models, all NN models are first trained, and then tested in CPU mode using the Matlab Neural Network Toolbox.


From Table \ref{clustering2},
DLTF TF largely surpasses AE-1 in term of both ACC and NMI, especially considering the fact that the two have the identical amount of parameters and that ReLU is also known to introduce sparsity. It is further interesting to find that DLTF TF even achieves more favorable clustering quality than AE-2, with lower complexity and half running time in practice. Those results demonstrates that DLTF TF is both efficient and effective for clustering. 

The state-of-the-art deep clustering model DEC achieves better ACC/NMI results than DLTF TF and else. However, it involves much more parameters and higher complexity: the testing time of DEC is thus five times that of DLTF TF. Moreover, DEC is the only approach here that is specifically optimized for clustering, while DLTF and others all lead to general-purpose unsupervised features. As a minimal-complexity and general-purpose feature extraction way, DLTF TF provides a \textit{complementary effort} to dedicatedly-trained and high-complexity deep clustering models. We also recognize that, to obtain stronger clustering performance, DLTF can be easily integrated with other regularizations, e.g. graph Laplacian \cite{zheng2011graph}, as well as can be further optimized in a (clustering) task-driven way, e.g., using bi-level optimization \cite{mairal2012task}. 








\subsection{Experiments on Image Denoising}

\begin{table*}[htp!]
\caption{The PSNR comparison (dB) between the KSVD denoising \cite{KSVD} and DLTF methods of different $k$ values, under various $\sigma$ levels, for the \textit{House} image.}
\centering{
\setlength{\tabcolsep}{15pt}
\resizebox{0.75\textwidth}{!}{
\begin{tabular}{| c | c | c | c | c | c|}
\hline
\multicolumn{2}{|c|}{$\sigma$} & 20 & 25 & 30 & 40  \\
\hline
\multicolumn{2}{|c|}{Input} & 22.09 & 20.07  &  18.56 & 16.09 \\
\hline
\multicolumn{2}{|c|}{KSVD} & 31.44 & 30.47  &  29.58 & 27.34 \\
\hline
\multirow{4}{*}{DLTF} & $k$ = 1 & 30.04 & 29.25 & 28.42 &  26.78  \\
\cline{2-6}
& $k$ = 2 & 28.56 & 27.98 & 27.59 &  25.73  \\
\cline{2-6}
& $k$ = 4 & 27.97 & 27.05 & 26.85 &  25.54  \\
\cline{2-6}
& $k$ = 6 & 24.06 & 23.97 & 23.54 &  25.57  \\
\cline{2-6}
& $k$ = 8 & 21.63 & 21.61 & 21.31 &  22.75  \\
\hline
\end{tabular}}}
\label{noise}
\end{table*}

\begin{table*}[htp!]
\caption{The mAP comparison on the CIFAR-10 dataset for unsupervised hashing, at different code lengths $m$.}
\centering{
\setlength{\tabcolsep}{15pt}
\resizebox{0.75\textwidth}{!}{
\begin{tabular}{| c || c | c | c | c | c|}
\hline
Methods  & LSH & SH & SpH & SparseH & DLTF \\
\hline
\hline
$m$ = 32 &  0.166 & 0.089 &  0.145 & 0.149 & 0.148  \\
\hline
$m$ = 64  &  0.159 & 0.121 &  0.189 &  0.234 & 0.223 \\
\hline
$m$ = 128 & 0.268 & 0.165 & 0.232 &  0.312 &  0.296 \\
\hline
$m$ = 256  & 0.332 &  0.201 & 0.255 & 0.386 & 0.355 \\
\hline
\end{tabular}}}
\label{map}
\end{table*}

Sparse coding has found successful applications in image denoising~\cite{KSVD,vincent2010stacked,Liu_LWLWH_IJCAI2018}, making it an appealing benchmark problem to evaluate the effectiveness of thresholded feature and DLTF too. The classical KSVD algorithm was designed to perform patch-level image denoising \cite{KSVD}: by first dividing the noisy image into patches, from which a patch-level dictionary $W$ tailored to the input image is learned; OMP is then performed on each patch w.r.t. $W$ to compute the sparse codes $z$, in order to obtain the denoised patch $W^T z$; patches are eventually arranged back to the denoised image with overlapped regions averaged. Our DLTF-based image denoising approach follows the same pipeline except for two changes: (1) replacing the KSVD dictionary learning with DLTF; (2) replacing the patch-wise sparse codes solved by OMP with the thresholded feature (\ref{feature}).

We use the dictionary of $m = 256, n = 64$, designed to handle image patches of size $8 \times 8$ pixels. The \textit{House} image is used, with the Gaussian noise standard deviations ranging in [20, 25, 30, 40]. We sample a set of 5,000 $8 \times 8$ patches from the noisy image itself for dictionary learning in both KSVD and DLTF. Table \ref{noise} compares the KSVD results, and DLTF denoising results at $k$ chosen at 1, 2, 4, 6, and 8, under various $\sigma$ levels. It is encouraging to observe that DLTF is able to achieve quite close performance to KSVD at all noise levels, with a gap around or less than 1 dB. As for the choice of $k$, DLTF shows a strong preference of DLTF for the smallest $k$ = 1.
That agrees with the previous observation \cite{KSVD} that a self-trained dictionary will lead to high sparsity (e.g., 1-2 nonzero coefficients per patch), and is different from clustering experiments where a medium $k$ value is preferred by DLTF. Figure \ref{noise} displays a group of visual results at $\sigma$ = 20, where the DLTF denoising results at $k$ = 1, 2 show impressive capability in preserving edges and suppressing ringing artifacts. 



We also compare the testing time of KSVD and DLTF denoising, i.e., the average running time of computing OMP and thresholded features, respectively. For the \textit{House} image, our plain Matlab implementation of computing (\ref{feature}) costs 0.83s, while the state-of-the-art OMP implementation OMP-Box v10 \footnote{{\url{http://www.cs.technion.ac.il/~ronrubin/software.html}}} costs 1.69s in the same testing environment. However, it is noteworthy that OMP-Box v10 relies on the specifically-optimized Batch OMP algorithm (\cite{rubinstein2008efficient}), as well as optimized \underline{MEX functions written in C}, making the time comparison actually \underline{unfair and to our disadvantage}. Our future work plans to develop efficient C implementations for computing (\ref{feature}) for large-scale signals as well, 

\subsection{Experiments on Unsupervised Hashing}

Hashing is one of the popular solutions for approximate nearest neighbor search because of its low storage cost and fast retrieval speed. A lot of methods have been proposed to learn effective hash function. Hash codes of the same cluster are similar to each other while the hash codes in different clusters are dissimilar. In this section, we evaluate the proposed DLTF approach on the task of unsupervised hashing, inspired by the previous success of sparse hashing \cite{zhu2013sparse}, as well as the nonzero support itself as a natural binary feature. It is also more challenging as unsupervised hashing relies on only unlabeled data to generate binary hashing codes. We choose local sensitive hashing (LSH) \cite{charikar2002similarity}, spectral hashing (SH) \cite{weiss2009spectral}, and spherical hashing (SpH) \cite{heo2012spherical} as three classical baselines, following their default settings. We also implement a sparse coding-based hashing (SparseH) approach, that follows the pipeline of \cite{zhu2013sparse}
: a dictionary is learned on the training set, which is then used to solve sparse codes of the query images iteratively; sparse codes are binarized by encoding non-zero entries to ones and zeros elsewhere. Finally, DLTF is adapted for hashing by solving the dictionary $W$ from (\ref{dl}), computing $\bar{\boldsymbol{z}}$ via (\ref{feature}), and using $\text{sgn}(\bar{\boldsymbol{z}})$ as binary codes. We choose $k = \frac{m}{2}$ \footnote{$k$ = $m/2$ is chosen to meet the \{0,1\} bit balance, a desirable property for code efficiency. While the sufficient recovery conditions may not be satisfied for the large $k$, DLTF achieves good empirical performance.}at each code length $m$. $\lambda = 0.1$ and $\theta = 0.01$ are fixed.

We test on the CIFAR-10 \cite{krizhevsky2009learning} benchmark, which has been widely used to evaluate both supervised and unsupervised hashing methods. We use GIST descriptors $\in \R^{512}$ to represent each image, and discard the label. A query set is formed by randomly choosing 1,000 samples, and a non-overlapping training set is constructed using the rest. We vary the hashing code length $m$ from 32 bits to 256 bits, to evaluate the performance of all methods on compact codes and relatively long codes. The \textit{mean average precision} (mAP) is evaluated at different numbers of bits. As seen from Table \ref{map}, while SparseH maintains the best performance among all in most cases (except being outperformed by LSH at $m$ = 32), DLTF produces comparable mAPs and usually ranks only next to SparseH. Notice that the mAP difference between SparseH and DLTF is minimal at small $m$ = 32 or 64. 

Moreover, hashing applications emphasize high query efficiency and low latency. SparseH inevitably suffers from the heavy computational overhead of iterative sparse coding inference. DLTF achieves comparable results with one to two orders of magnitude less testing time (similar to the previous time comparison between DLTF and KSVD baselines). 

\subsection{Exploring A Deeper Potential}

Recently, deep learning~\cite{Xu_XLWRC_2018,Ranjan_RBZXGLNCCC_2018,He_HZRS_CVPR2016,Simonyan_SZ_2014,Sermanet_SEZMFL_ICLR2014,Yan_YLYP_PIEEE1998} has achieved significant success on many computer vision tasks such as image classification, semantic segmentation and object detection. Despite the superior performance of deep learning-based methods, most networks do not take advantage of successful conventional ideas. We explore a deeper potential in this section. 

The proposed DLTF approach is mainly for unsupervised learning task, while recognizing a potential link between DLTF and (supervised) deep models. 
Eqn. (\ref{feature}) can be viewed as a fully-connected layer followed by a non-linearity function, one of the standard building blocks in existing deep models\footnote{Here max$_k$ could be thought either as a generalization of max pooling without reducing dimension, or a locality-aware neuron.}. In that way, DLTF can also be viewed as a variant of AE, where max$_k$ acts as the nonlinear function to promote the sparsity in hidden layer activations. Its objective is to recover the sparse support besides maintaining faithful reconstruction. Then a natural question arises: will DLTF be a competitive alternative in layer-wise unsupervised pre-training of deep models?

\begin{table}[htp!]
\caption{The error rate comparison for MNIST classification, using different pre-training strategies.}
\centering{
\setlength{\tabcolsep}{7pt}
\resizebox{0.48\textwidth}{!}{
\begin{tabular}{| c || c | c | c |}
\hline
Pre-Training  & Random (No)  & AE  & DLTF  \\
\hline
\hline
Model 1 &  1.62\% & 1.08\% &  1.04\%  \\
\hline
Model 2  &  1.26\% & 0.89\% &  0.81\%  \\
\hline
\end{tabular}}}
\label{AE}
\end{table}

We further conduct preliminary experiments on comparing AE and DLTF in \textit{pre-training} neural networks for classifying the MNIST dataset. More specifically, we construct two fully connected network models for the 10-class MNIST classification benchmark (note the different setting with the clustering experiments in Section 4.2). Model 1 takes 784-dimensional inputs, followed by a 1000-dimensional hidden layer and the softmax loss. Model 2 has one additional 1000-dimensional hidden layer appended before the loss function. For Model 1, we perform DLTF with $n = 784, m = 1000$ and choose $k = 100$, to learn the dictionary as the initialization for its hidden layer. For Model 2, we perform DLTF with $n = 784, m = 1,000, k = 200$ for the first hidden layer; we then fix the first layer, and perform DLTF on the first layer outputs with $n = 1,000, m = 1,000, k = 100$, for the second layer. We compare DLTF pre-training with random initializations (no pre-training), and classical AE pre-training. All models are then tuned from end to end, and dropout with a ratio of 0.5 is applied to all fully-connected layers during fine-tuning. Table \ref{AE} compares the error rates, where DLTF shows an advantage.

Table \ref{AE} potentially implies that identifying the correct parameter subspace is a more promising goal for layer-wise pre-training, than minimizing the MSE. The hypothesis, if validated further, could lead to new insights of pre-training or even training deep models. We will conduct more experiments to verify if DLTF pre-training can benefit the training of more general deep models. 

Despite being preliminary, the results suggest that identifying the correct (sparse) parameter subspace (i.e., the nonzero support) may be a more promising goal for layer-wise pre-training, than solely minimizing the reconstruction error. It is interesting to further explored that DLTF pre-training can benefit more general deep models

\section{Conclusion and Future Work}
In this paper, motivated by support recovery theoretical guarantees, 
we propose a novel approach to learn a dictionary which is optimized for applying the thresholded feature. The competitive performance and superior efficiency of the proposed approach are extensively studied in both synthetic simulations and real-data experiments. In future work, we seek more elaborating formulations of DLTF. For example, \cite{lin2015optimized} suggested that minimizing $||W^\top W - I||_\infty$ could suppress $\mu_W$ better, 
although minimizing the former term is also accompanied with higher complexity. 


%
\appendices

\section{Proof of Theorem 1: The Weak Recovery Guarantee}\label{proof_Theorem1}
\begin{proof} Denote $\boldsymbol{y}=W^\top  \boldsymbol{x} = W^\top W \boldsymbol{z}$. Then for  $0 \leq i \leq k$:
\begin{equation}
 \begin{aligned}
   |y_i| & = |z_i+\sum_{j=1,j \neq i}^{k} \langle \boldsymbol{w}_i,\boldsymbol{w}_j \rangle z_j| \\
   & \geq |z_i| - \mu_W\sum_{j=1,j \neq i}^{k} |z_j| \geq |z_{k}| - k \mu_W |z_1|  
  \end{aligned}
\end{equation}
On the other hand, for $i > k$:
\begin{equation}
 \begin{aligned}
   \underset{i>k}{\text{max }} \{|y_i|\} &= \underset{i>k}{\text{max }} \left\{ |\sum_{j=1}^{k} \langle \boldsymbol{w}_i,\boldsymbol{w}_j \rangle z_j| \right\}
   &\leq k\mu_W |z_1|  
  \end{aligned}
\end{equation}
So if $2 k\mu_W |z_1|\leq |z_{k}|$, the first \emph{k} entries of $\boldsymbol{y}$ are guaranteed to have greater magnitudes than the rest, and thus will be correctly identified. 
\end{proof} 

Theorem 1 reveals a sufficient condition on the required number of samples to guarantee the selection consistency, and a similar conclusion could be found in \cite{makhzani2013k}. Following a similar analysis, additional results with regard to the decay rate of nonzeros may be obtained. Moreover, based on the random matrix theory \cite{edelman2005random} and that $\mu_W \le 1$ due to the normalization of $W$, $\mu_W$ will decay to 0 with the rate $\frac{\log(n)}{m}$  if $W$ is a sub-Gaussian random matrix. Therefore, Theorem 1 is to reveal the required \# of samples to guarantee the selection consistency.

\section{Proof of Theorem 2: The Strong Recovery Guarantee}\label{proof_Theorem2}
\begin{proof}
Define $\Omega_z=\text{supp}(\boldsymbol{z}), \Omega_{\bar{z}}=\text{supp}(\bar{\boldsymbol{z}})$, $S = \Omega_z\cup\Omega_{\bar{z}}$. Let $[\boldsymbol{z}]_S$ denotes the subvector indexed by the set $S$. First we have
\begin{equation}
  \begin{aligned}
    \| \bar{\boldsymbol{z}}-W^\top\boldsymbol{x} \|^2 = & \| \bar{\boldsymbol{z}}-\boldsymbol{z} \|^2  +\| \boldsymbol{z}-W^\top\boldsymbol{x} \|^2 \\
    & +2\langle \bar{\boldsymbol{z}}-\boldsymbol{z}, \boldsymbol{z}-W^\top\boldsymbol{x} \rangle.
  \end{aligned}
\end{equation}
Since $\| \bar{\boldsymbol{z}}-W^\top\boldsymbol{x} \|^2\leq\| \boldsymbol{z}-W^\top\boldsymbol{x} \|^2$ due to the projection property, we have
\begin{equation}
 \begin{aligned}
    \| \bar{\boldsymbol{z}}-\boldsymbol{z} \|^2 & \leq 2\langle \bar{\boldsymbol{z}}-\boldsymbol{z}, W^\top\boldsymbol{x}-\boldsymbol{z} \rangle \\
    & =2\langle \bar{\boldsymbol{z}}-\boldsymbol{z}, [W^\top\boldsymbol{x}-\boldsymbol{z}]_S \rangle \\
    & \leq 2 \| \bar{\boldsymbol{z}}-\boldsymbol{z} \| \| [W^\top\boldsymbol{x}-\boldsymbol{z}]_S \|
 \end{aligned}
\end{equation}
It follows
\begin{align*}
  & \| \bar{\boldsymbol{z}}-\boldsymbol{z} \| \leq 2\| [W^\top\boldsymbol{x}-\boldsymbol{z}]_S \|\\
= & 2 \| [W^\top\boldsymbol{x}-\boldsymbol{z}+W^\top(W\boldsymbol{z}-\boldsymbol{x})-W^\top(W\boldsymbol{z}-\boldsymbol{x})]_S \| \\
\leq & 2 \| [W^\top\boldsymbol{x}-\boldsymbol{z}+W^\top(W\boldsymbol{z}-\boldsymbol{x})]_S \| + 2\| [W^\top(W\boldsymbol{z}-\boldsymbol{x})]_S \| \\
= & 2\| \boldsymbol{z} - [W^\top\boldsymbol{x} + W^\top(W\boldsymbol{z}-\boldsymbol{x})]_S\| + 2\| [W^\top(W\boldsymbol{z}-\boldsymbol{x})]_S \| \\
= & 2 \| \boldsymbol{z} - [W^\top W\boldsymbol{z}]_S\| + 2\| [W^\top\boldsymbol{e}]_S \|.
\end{align*}   
Using Assumption 1, we have 
\begin{align*}
& \| \boldsymbol{z}-[W^\top W\boldsymbol{z}]_S \|^2 \\
= & \| \boldsymbol{z} \|^2 + \| [W^\top W\boldsymbol{z}]_S \|^2 - 2\langle\boldsymbol{z},[W^\top W\boldsymbol{z}]_S\rangle \\
\leq & \| \boldsymbol{z} \|^2 +((1+\delta_W)-2)\langle \boldsymbol{z},[W^\top W \boldsymbol{z}]_S\rangle \\
\leq & \| \boldsymbol{z} \|^2 -(1-\delta_W)^2\| \boldsymbol{z} \|^2
= (2 \delta_W -  \delta_W^2) \| \boldsymbol{z} \|^2
\end{align*}
$\delta_W \in (0, 1 - \frac{\sqrt{3}}{2})$ is required to ensure $2 \delta_W -  \delta_W^2 > 0$. Also considering $|S| \le 2k$, we have
\begin{align}
  \| \bar{\boldsymbol{z}}-\boldsymbol{z} \| \leq 2\sqrt{2 \delta_W -  \delta_W^2} \| \boldsymbol{z} \| + 2\left\| \text{max}_{\text{\emph{2k}}}(W^\top \boldsymbol{e}) \right\| \label{eq:recur}
\end{align}
Therefore, if the smallest nonzero element $z_k$ of $\boldsymbol{z}$ is no less than the right side of (\ref{eq:recur}), $\bar{\boldsymbol{z}}$ and $\boldsymbol{z}$ must have the same support set. It completes the proof.
\end{proof}

The Restricted Isometry Property (RIP) has been a fundamental concept in sparse recovery \cite{candes2005decoding, yang2016benefits}.  Both RIP and mutual coherence require $W$ to behave like an orthonormal system. A recent result \cite{wang2015linear} reveals both to be special forms of a more generalized sufficient condition for sparse recovery.  

\section{Proof of Theorem 3}

To use Algorithm~1 to solve the proximal mapping problem~(2) for any $\boldsymbol{c} \in \R^m$, we firstly turn the input $\boldsymbol{c}$ into positive vector $\boldsymbol{c}'$ by taking its absolute values (i.e. $\boldsymbol{c}'=|\boldsymbol{c}|$). The solution to the original problem can be obtained by using the sign of $\boldsymbol{c}$ because we know that the sign of the proximal mapping $prox^{k',2}_{\gamma}(\boldsymbol{c})$ is the same as of $\boldsymbol{c}$. Secondly, we sort the elements in $\boldsymbol{c}'$, which will meet the input requirement of Algorithm~1. Then we can use Algorithm~1 to get the proximal mapping. The sorting procedure costs $O(m\log m)$ time, and the Algorithm~1 costs $O(m)$ time, so the total time complexity is $O(m\log m)$.

When the elements of $\boldsymbol{c}$ are in increasing order, Lemma 5 converts the problem (11) to:
\begin{gather}
\min_{\boldsymbol{q}}\, \| \boldsymbol{q}-\boldsymbol{c} \|^2 + \gamma \sum_{i=m-k'+1}^m q_i^2, \, \\
s.t.\, \, q_1 \leq q_2 \leq ... \leq q_m \notag
\end{gather}
Dropping some constant terms, the objective can be further written as:
\begin{gather}
\min_{\boldsymbol{q}} \sum_{j=1}^{m-k'} (q_j - c_j)^2 + \sum_{j=m-k'+1}^m (1+\gamma) (q_j - {1\over \lambda + 1}c_j)^2 \label{eq:order_obj2}\\
\,s.t.\, \, q_1 \leq q_2 \leq ... \leq q_m \notag
\end{gather}
which can be solved by Algorithm 1.

\section{Proof of Lemma 4}
\begin{proof}
We prove by contradiction. Suppose that $c_i < c_j$ and $\boldsymbol{q}^*_i > \boldsymbol{q}^*_j$, we will show that we can get a better solution $\mathbf{v}'$ by swapping $\boldsymbol{q}*_i$ and $\boldsymbol{q}^*_j$, which makes a contradiction. 

Firstly, we have $q^*_i \leq c_i, q^*_j \leq c_j$ since $\boldsymbol{q}^*$ is optimal. Let the objective value be $o_1$ before we swap $q^*_i$ and $q^*_j$, and be $o_2$ after swap.
From the objective definition, the objective value will change:
\begin{equation}
o_1 - o_2 = (c_i - q^*_i)^2 + (c_j - q^*_j )^2 - (c_i - q^*_j)^2 - (c_j - q^*_i)^2
\end{equation}
Let $a=(c_i - q^*_i), b= (c_i - q^*_j), d=c_i - q^*_i + c_j - q^*_j$, then
\begin{align*}
& o_1 - o_2\\
= & a^2+(d-a)^2 - b^2 - (d-b)^2\\
=& 2(a^2 - b^2 - d(a-b)) \\
=& 2(a-b)(a+b-d)
\end{align*}
where $a-b =q^*_j - q^*_i < 0$, and $a+b-d = c_i - c_j <0$, so
$o_1 > o_2$, which contradicts the assumption that $o_1$ is optimal. It completes the proof.
\end{proof}

\section{Proof of Lemma 5 and Lemma 6}

\begin{proof}

We first prove Lemma 6 by contradiction. Assume $\u_j > \u_{j+1}$ and $\x_j^*\neq \x^*_{j+1}$, then we have $\x_j^*<\x^*_{j+1}$ because we have the constraint (11). Next the contradiction will be shown below by illuminating all six scenarios. We essentially will show there always exists a solution pair $(\x'_j, \x'_{j+1})$ satisfying $\x'_j = \x'_{j+1} \in [\x_j^*, \x^*_{j+1}]$ which gives a better solution than $(\x_j^*, \x^*_{j+1})$ in every scenario.
\begin{itemize}
\item $\x^*_j \leq \u_{j+1} <  \u_j \leq \x^*_{j+1}$: We can choose $\x'_j = \x'_{j+1} = \u_j$;
\item $\u_{j+1} \leq \x^*_j < \x^*_{j+1} \leq \u_j$: We can choose $\x'_j = \x'_{j+1} = \x^*_j$;
\item $\x^*_j  < \u_{j+1} \leq \x^*_{j+1} \leq \u_j$: We can choose $\x'_j = \x'_{j+1} = \x^*_{j+1}$;
\item $\u_{j+1} \leq \x^*_j < \u_j \leq \x^*_{j+1}$: We can choose $\x'_j = \x'_{j+1} = \u_{j}$;
\item $\u_{j+1} < \u_j \leq \x^*_j  < \x^*_{j+1}$: We can choose $\x'_j = \x'_{j+1} = \x^*_{j}$;
\item $\x^*_j  < \x^*_{j+1} \leq \u_{j+1} < \u_j $: We can choose $\x'_j = \x'_{j+1} = \x^*_{j+1}$.
\end{itemize}
It completes the proof of Lemma 6.

Now let us merge two successive variables $\x_j$ and $\x_{j+1}$ if we find $\u_j > \u_{j+1}$. From Lemma 6, we know $\x_j$ should be equal to $\x_{j+1}$. Introduce a new variable $\x_{j\vee j+1}$ to denote the value of $\x_j$ and $\x_{j+1}$. It follows that the original problem in Lemma 5 is equivalent to solving
\begin{align}
& \min_{\x_1 \leq \cdots \leq \x_{j-1} \leq \x_{j} = \x_{j+1} \leq \x_{j+2} \leq \cdots \leq \x_J}\quad \sum_{j=1}^{J}{\mathbf{t}_j}(\x_j-\u_j)^2 \label{eq:proof:prox:0} 
\\
& \Leftrightarrow \hspace{1.5cm} \min_{\x_1 \leq \cdots \leq \x_{j-1} \leq \x_{j\vee j+1} \leq \x_{j+2} \leq \cdots \leq \x_J}\quad \notag \\ 
& \hspace{1.5cm} {\mathbf{t}_j+\mathbf{t}_{j+1} } \left(\x_{j\vee j+1} - {\mathbf{t}_j \u_j + \mathbf{t}_{j+1} \u_{j+1} \over \mathbf{t}_j+ \mathbf{t}_{j+1}} \right)^2 \notag \\
& \hspace{2.5cm} + \sum_{i\notin \{j,j+1\}} {\mathbf{t}_i }(\x_i-\u_i)^2 \notag \\
& \Leftrightarrow \hspace{1.5cm} \min_{\x_1 \leq \cdots \leq \x_{j-1} \leq \x_{j\vee j+1} \leq \x_{j+2} \leq \cdots \leq \x_J}\quad \notag \\
& \hspace{0.8cm} {\mathbf{t}_{j\vee j+1} } \left(\x_{j\vee j+1} - \u_{j\vee j+1} \right)^2 + \sum_{i\notin \{j,j+1\}} {\mathbf{t}_i }(\x_i-\u_i)^2
\label{eq:proof:prox:1} 
\end{align}
where in the last line 
$$
\mathbf{t}_{j\vee j+1}= \mathbf{t}_j+\mathbf{t}_{j+1} \quad \u_{j\vee j+1} = {\mathbf{t}_j \u_j + \mathbf{t}_{j+1} \u_{j+1} \over \mathbf{t}_j+ \mathbf{t}_{j+1}}.
$$
Therefore, if we define 
\begin{align*}
\u' & = [\u_1, \cdots, \u_{j-1}, \u_{j\vee j+1}, \u_{j+2}, \cdots, \u_J]^\top \\
\mathbf{t}' & = [\mathbf{t}_1, \cdots, \mathbf{t}_{j-1}, \mathbf{t}_{j\vee j+1}, \mathbf{t}_{j+2}, \cdots, \mathbf{t}_J]^\top,
\end{align*}
then we reduce the problem dimension to $J-1$. It should be noticed that we do not need to solve the reduced problem from the beginning, since $\u'_{1:j-1}=\u_{1:j-1}$. By this way, Algorithm~1 can be completed in linear time $O(m)$.

Then the solution to the original one \eqref{eq:proof:prox:0} can be recovered by extending the $j$th element of Reduce($\u', \mathbf{t}', j-1$) to the $(j+1)$th element. We can keep checking if there are any two successive components in $\u'$ disobeying the nondecreasing monotonicity. As long as we find one pair, we can reduce the original problem by one dimension. Therefore, this problem can be recursively solved by the subroutine ``Reduce'' in Algorithm~1. Then it completes the proof of Lemma 5.
\end{proof}





\ifCLASSOPTIONcaptionsoff
  \newpage
\fi

{
\bibliographystyle{ieee}
\bibliography{egbib_tsp}
}

\begin{IEEEbiography}[{\includegraphics[width=1in,height=1.25in,clip,keepaspectratio]{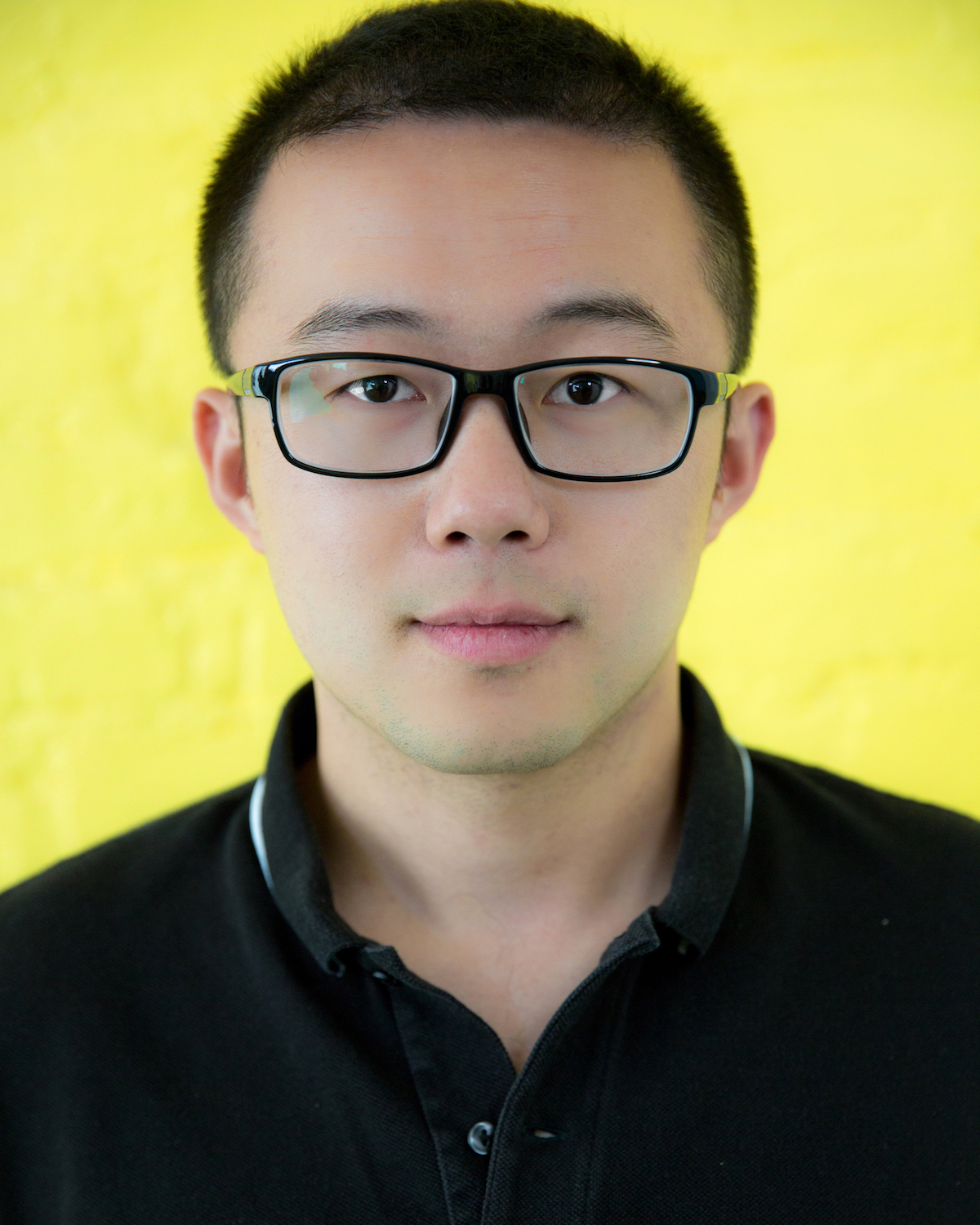}}]{Hongyu Xu}
received the B.E. degree from the University of Science and Technology of China in 2012 and the M.S. degree from University of Maryland, College Park in 2017. He is currently a research assistant in the Institute for Advanced Computer Studies at the University of Maryland, College Park, advised by Prof. Rama Chellappa. He is a former research intern with Snap Research (summer, fall 2017) and Palo Alto Research Center (PARC) (summer 2014). His research interests include object detection, deep learning, dictionary learning, face recognition, and domain adaptation.
\end{IEEEbiography}

\begin{IEEEbiography}[{\includegraphics[width=1in,height=1.25in,clip,keepaspectratio]{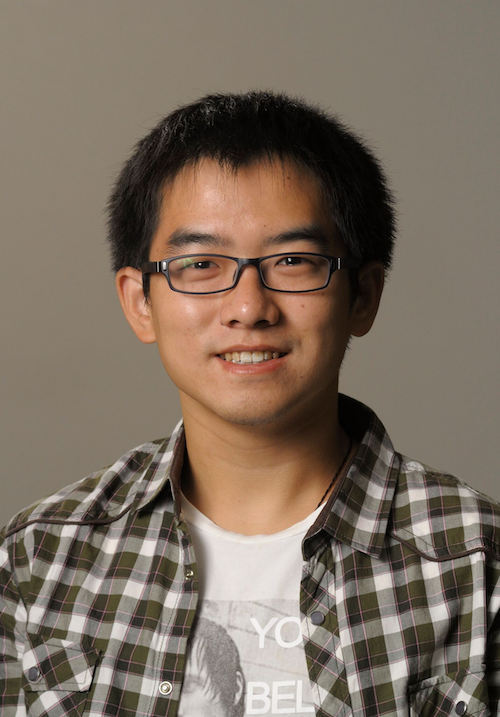}}]{Zhangyang Wang} is an Assistant Professor of Computer Science and Engineering (CSE), at the Texas A\&M University (TAMU). During 2012-2016, he was a Ph.D. student in the Electrical and Computer Engineering (ECE) Department, at the University of Illinois at Urbana-Champaign (UIUC), working with Professor Thomas S. Huang. Prior to that, he obtained the B.E. degree at the University of Science and Technology of China (USTC), in 2012. He is a former research intern with Microsoft Research (summer 2015), Adobe Research (summer 2014), and US Army Research Lab (summer 2013). Dr. Wang’s research has been addressing machine learning, computer vision and multimedia signal processing problems, as well as their interdisciplinary applications, using advanced feature learning and optimization techniques. He has co-authored over 60 papers, and published several books and chapters. He has been granted 3 patents, and has received around 20 research awards and scholarships. 
\end{IEEEbiography}

\begin{IEEEbiography}[{\includegraphics[width=1in,height=1.25in,clip,keepaspectratio]{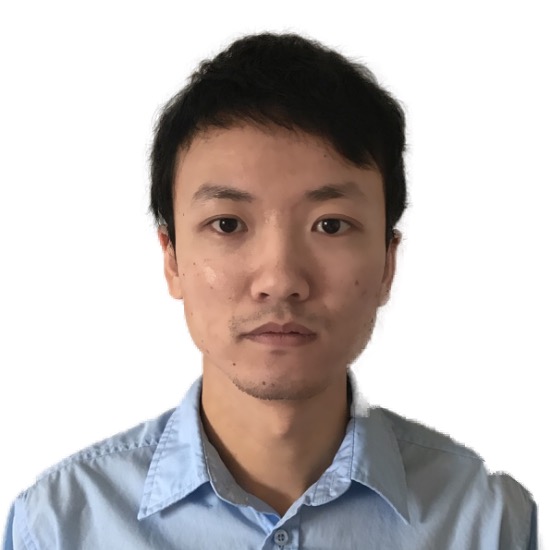}}]{Haichuan Yang} is currently a Ph.D. student (since 2015) at Department of Computer Science, University of Rochester, Rochester, N.Y. Before that, he received the bachelor's degree in 2012 in Software Engineering from Sun Yat-sen University, Guangzhou, China and the master's degree in 2015 in Computer Science and Engineering from Beihang University, Beijing, China. His research interests focus on deep learning and machine learning, particularly in deep neural network compression, sparse optimization and reinforcement learning.
\end{IEEEbiography}

\begin{IEEEbiography}[{\includegraphics[width=1in,height=1.25in,clip,keepaspectratio]{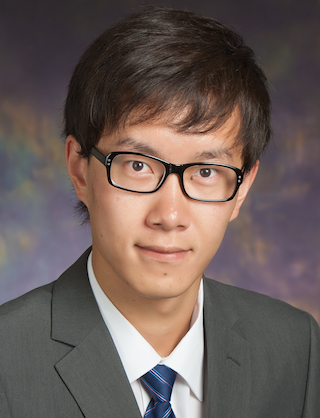}}]{Ding Liu} received the B.S. degree from the Chinese University of Hong Kong, Hong Kong, in 2012, and the M.S. and Ph.D. degrees from the University of Illinois at Urbana-Champaign, USA, in 2014 and 2018, respectively. His research expertise encompasses image restoration and image enhancement. He has research interests in the broad area of computer vision, image processing and deep learning. 
\end{IEEEbiography}

\begin{IEEEbiography}[{\includegraphics[width=1in,height=1.25in,clip,keepaspectratio]{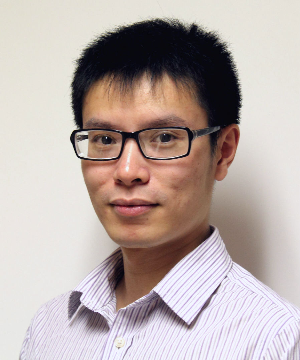}}]{Ji Liu} is currently an assistant professor in Computer Science, Electrical Computer Engineering, and Goergen Institute for Data Science at University of Rochester (UR). He received his Ph.D., Masters, and B.S. degrees from University of Wisconsin-Madison, Arizona State University, and University of Science and Technology of China respectively. His research interests cover a broad scope of machine learning, optimization, and their applications in other areas such as healthcare, bioinformatics, computer vision, game AI, and many other data analytics related areas. His recent research focus is on reinforcement learning, structural model estimation, asynchronous parallel algorithms, decentralized algorithms, sparse learning (compressed sensing) theory and algorithm, healthcare, bioinformatics, etc. He founded the machine learning and optimization group at UR and published more than 40 papers in top conferences and journals. He won the award of Best Paper honorable mention at SIGKDD 2010, the award of Facebook Best Student Paper award at UAI 2015, the IBM faculty award 2017, and the MIT TR35 award China. 
\end{IEEEbiography}

\end{document}